\documentclass[a4paper,fleqn]{cas-dc}

\usepackage[numbers]{natbib}
\usepackage[ruled,vlined]{algorithm2e}
\usepackage{rotating}
\usepackage{subfigure}

\def\tsc#1{\csdef{#1}{\textsc{\lowercase{#1}}\xspace}}
\tsc{WGM}
\tsc{QE}

\begin{document}
\let\WriteBookmarks\relax
\def\floatpagepagefraction{1}
\def\textpagefraction{.001}

\shorttitle{Reinforced Hybrid GA for the TSP}    

\shortauthors{Zheng et al.}  

\title [mode = title]{Reinforced Hybrid Genetic Algorithm for the Traveling Salesman Problem}  

\author[1,2]{Jiongzhi Zheng}

\credit{Conceptualization of this study, Methodology, Software, Writing and revision}

\address[1]{School of Computer Science, Huazhong University of Science and Technology, China}

\author[1]{Jialun Zhong}
\credit{Writing and revision}
\author[1]{Menglei Chen}
\credit{Writing and revision}

\author[1]{Kun He}
\cormark[1]
\cortext[cor1]{Corresponding author. Email: brooklet60@hust.edu.cn}
\credit{Conceptualization of this study, Methodology, Supervision, Writing and revision}

\begin{abstract}
In this paper, we propose a new method called the Reinforced Hybrid Genetic Algorithm (RHGA) for solving the famous NP-hard Traveling Salesman Problem (TSP). Specifically, we combine reinforcement learning with the well-known Edge Assembly Crossover genetic algorithm (EAX-GA) and the Lin-Kernighan-Helsgaun (LKH) local search heuristic. In the hybrid algorithm, LKH can help EAX-GA improve the population by its effective local search, and EAX-GA can help LKH escape from local optima by providing high-quality and diverse initial solutions. We restrict that there is only one special individual among the population in EAX-GA that can be improved by LKH. Such a mechanism can prevent the population diversity, efficiency, and algorithm performance from declining due to the redundant calling of LKH upon the population. As a result, our proposed hybrid mechanism can help EAX-GA and LKH boost each other's performance without reducing the convergence rate of the population. The reinforcement learning technique based on Q-learning further promotes the hybrid genetic algorithm. Experimental results on 138 well-known and widely used TSP benchmarks with the number of cities ranging from 1,000 to 85,900 demonstrate the excellent performance of RHGA.
\end{abstract}

\begin{keywords}
Combinatorial optimization \sep Traveling salesman problem \sep Hybrid genetic algorithm \sep Local search \sep Reinforcement learning
\end{keywords}

\maketitle

\sloppy{}

\section{Introduction}
\label{sec_intro}
Given a complete, undirected graph $G = (V, E)$, where $V = \{1, 2, ..., n\}$ denotes the set of $n$ cities and $E = \{(i,j) | i,j \in V\}$ denotes the set of all pairwise edges, $d(i,j)$ represents the distance (cost) of edge $(i,j)$, i.e., the distance of traveling from city $i$ to city $j$, the Traveling Salesman Problem (TSP) aims to find a Hamiltonian cycle represented by a permutation $(s_1, s_2, ..., s_n)$ of cities $\{1, 2, ..., n\}$ that minimizes the total distance, i.e., $d(s_1,s_2) + d(s_2,s_3) + ... + d(s_{n-1},s_n) + d(s_n,s_1)$. The TSP is one of the most famous and well-studied NP-hard combinatorial optimization problems, which is very easy to understand but very difficult to solve to the optimality. Over the years, the TSP has become a touchstone in the field of the combinatorial optimization.

Typical methods for solving the TSP can be categorized into exact algorithms, approximation algorithms, and heuristics. The exact algorithms may be prohibitive for large instances, and the approximation algorithms may suffer from weak optimal guarantees or empirical performance~\cite{Khalil2017}. Heuristics are known to be the most efficient and effective approaches for solving the TSP. Two of the state-of-the-art heuristics are the Lin-Kernighan-Helsgaun (LKH) local search algorithm~\cite{Helsgaun2000} and the Edge Assembly Crossover genetic algorithm (EAX-GA)~\cite{Nagata2013}. Both of them provide the best-known solutions on many TSP benchmark instances.

As two representative heuristic algorithms, both LKH and EAX-GA have advantages and disadvantages. For example, EAX-GA is very efficient and powerful in solving TSP instances with tens to hundreds thousands of cities, providing the best-known solutions of the six famous instances with 100,000 to 200,000 cities in the Art TSP benchmarks\footnote{http://www.math.uwaterloo.ca/tsp/data/art/index.html}. But EAX-GA is hard to scale to super large instances, such as TSP instances with millions of cities, since the convergence of the population is too time-consuming. As an efficient local search algorithm, LKH can yield near-optimal solutions faster than EAX-GA does. It is also suitable for TSP instances with various scales, especially for super large instances, providing the best-known solution of the famous World TSP instance with 1,904,711 cities\footnote{http://www.math.uwaterloo.ca/tsp/world/index.html}. However, LKH is not as good as EAX-GA in solving the TSPs with 10,000 to 200,000 cities, since the population can help EAX-GA explore the solution space better than LKH does for instances with such large scales. 
Based on these characteristics, a straightforward idea is proposed spontaneously. That is, whether there is a reasonable way to combine EAX-GA with LKH and make use of their complementary, so as to help them boost each other.

There have been related studies trying to combine EAX-GA with LKH or its predecessor, the LK heuristic~\cite{Lin1973}. For example, Tsai et al.~\cite{Tsai2004} propose to combine the earliest version of EAX-GA~\cite{Nagata1997} with LK. Their proposed algorithm HeSEA reports better results than EAX-GA, LK, and LKH in solving TSP instances with at most 15,112 cities. However, HeSEA follows the similar hybrid mechanism of many other hybrid algorithms~\cite{Ulder1990,Freisleben1996,Merz1997,Nguyen2007,Whitley2010,Wang2014,Rashid2017,Ilin2020} that combine genetic algorithms with LK-based algorithms (or other local search methods like 2-opt). That is, applying the local search methods to optimize every individual in the current population or every surviving offspring generated. 
Such a mechanism has two disadvantages: 1) the population diversity will be broken because the local optimal solutions (of different tours) calculated by the same local search method are similar. 2) It is very time-consuming to frequently apply local search methods to calculate the local optimal solutions, as HeSEA~\cite{Tsai2004} shows worse efficiency and reports longer computation time than LK and LKH for large scale instances.

In addition, Kerschke et al.~\cite{Kerschke2018} propose to combine several TSP solvers including EAX-GA and LKH by a machine learning model based on supervised learning. The machine learning model can help their proposed hybrid solver select an appropriate solver to solve the input TSP instance. 
Such a hybrid is a simple combination of the TSP solvers, in which the solvers do not interact with each other.  
The solution of the hybrid solver (consists of only EAX-GA and LKH) is bounded by the better one of the solutions obtained by EAX-GA and LKH. 

In this paper, we propose a reinforcement learning~\cite{Sutton1998,Hu2020,Oladayo2022} based hybrid genetic algorithm for the TSP, called the Reinforced Hybrid Genetic Algorithm (RHGA), that combines EAX-GA with the LKH local search and further applies reinforcement learning to improve the performance. In the proposed RHGA, there is only one special individual (e.g., the first individual) in the population of EAX-GA that can be improved by the local search algorithm of LKH, because the redundant local search operations and local optimal solutions of LKH in the population may reduce the population diversity, the efficiency, as well as the solution quality of the genetic algorithm. Moreover, our proposed combination mechanism can make full use of the complementary of EAX-GA and LKH, and help them boost each other. As a result, the hybrid mechanism in our proposed algorithm can fix the aforementioned issues of the hybrid mechanisms in the existing hybrid genetic algorithms~\cite{Ulder1990,Freisleben1996,Merz1997,Tsai2004,Nguyen2007,Whitley2010,Wang2014,Rashid2017,Ilin2020} and hybrid solver~\cite{Kerschke2018} for the TSP and fits well with EAX-GA and LKH.

Moreover, we apply reinforcement learning~\cite{Sutton1998} to further improve the performance of the proposed hybrid genetic algorithm. We apply the technique proposed by Zheng et al.~\cite{Zheng2021} that employs reinforcement learning to learn an adaptive Q-value as a metric for evaluating the quality of the edges. We use the adaptive Q-value learned by the Q-learning algorithm~\cite{Sutton1998} to replace the important evaluation metrics of the edges used in the key steps in both LKH and EAX-GA. In this way, both LKH and EAX-GA can be enhanced by the reinforcement learning in our RHGA algorithm. Related studies of (reinforcement) learning based methods for the TSP are referred to Section~\ref{Sec_RL}, where we also describe the advantages of our reinforcement learning method over them.

The main contributions of this work are as follows:
\begin{itemize}
\item We propose a creative and distinctive hybrid mechanism to combine two of the state-of-the-art TSP heuristic algorithms, EAX-GA and LKH, through a special individual. In the proposed RHGA algorithm, EAX-GA and LKH can boost each other with the 
bridge of the special individual. 
\item We propose to combine reinforcement learning with the key steps of both EAX-GA and LKH to further improve the performance of the hybrid genetic algorithm. The adaptive Q-value learned by the Q-learning algorithm significantly outperforms the metrics used in EAX-GA and LKH for evaluating the quality of the edges. 
\item Our proposed techniques, including the hybrid mechanism of combining genetic algorithm with local search method and the method of combining reinforcement learning with the key search steps of heuristics, can be applied to solve various combinatorial optimization problems, such as variant problems of TSP, the vehicle routing problems and the graph coloring problems.
\item  Experimental results on 138 well-known and widely used TSP benchmarks with the number of cities ranging from 1,000 to 85,900 demonstrate the promising performance of our proposed algorithm.
\end{itemize}

\section{Related Works}
\label{Sec_RelatedWork}
For related works, we first introduce (reinforcement) learning based algorithms for solving the TSP, then briefly introduce the main ideas and approaches in the two state-of-the-art heuristic algorithms for solving the TSP, EAX-GA and LKH, which will also be incorporated into our proposed algorithm. For details of these two algorithms, we refer to~\cite{Nagata2013} and~\cite{Helsgaun2000}.

\subsection{Learning Based Algorithms for the TSP}
\label{Sec_RL}
(Reinforcement) learning based methods for the TSP can be divided into two categories. The first category is end-to-end methods~\cite{Bengio2021}, which are usually based on deep neural networks. When receiving an input TSP instance, they use the trained learning model to generate a solution directly. For example, Bello et al.~\cite{Bello2017} address TSP by using the actor-critic method to train a pointer network~\cite{Vinyals2015}. The S2V-DQN algorithm~\cite{Khalil2017} applies reinforcement learning to train a graph neural network so as to solve several combinatorial optimization problems, including minimum vertex cover, maximum cut, and TSP. Goh et al.~\cite{Goh2022} use an encoder based on a standard multi-headed transformer architecture and a Softmax or Sinkhorn~\cite{Cuturi2013,Emami2018} decoder to directly solve the TSP. These methods provide good innovations in the field of applying machine learning to solve combinatorial optimization problems. As for the performance, they can yield near-optimal or optimal solutions for the TSP instances with less than hundreds of cities. However, they are usually hard to scale to large instances (with more than thousands of cities) due to the complexity of deep neural networks.

Methods belonging to the second category combine (reinforcement) learning methods with traditional algorithms. Some of them use traditional algorithms as the core and frequently call the learning models to help explore the solution space or guide the search direction. For example, Liu and Zeng~\cite{Liu2009} employ reinforcement learning to construct mutation individuals in the previous version of EAX-GA~\cite{Nagata2006} and report better results than EAX-GA and LKH on instances with up to 2,392 cities. But the efficiency of their proposed algorithm is not as good as that of LKH. Costa et al.~\cite{Costa2020} and Sui et al.~\cite{Sui2021} use deep reinforcement learning to guide 2-opt and 3-opt local search operators, and report results on instances with no more than 500 cities. Other methods separate the learning models and traditional algorithms. They first apply (reinforcement) learning methods to yield initial solutions~\cite{Zhao2021} or some configuration information~\cite{Xin2021}, and then use traditional algorithms to find high-quality solutions followed the obtained initial solutions or information. Among them, the NeuroLKH algorithm~\cite{Xin2021} is one of the state-of-the-art, which uses a Sparse Graph Network with supervised learning to generate the candidate edges for LKH. It reports better or similar results compared with LKH in instances with less than 6,000 cities.

In summary, (reinforcement) learning based methods with deep neural networks for the TSP may suffer from the bottleneck of hardly solving large scale instances, and the combination of traditional reinforcement learning methods (training tables, not deep neural networks) with existing (heuristic) algorithms may reduce the efficiency of the algorithm. The reinforcement learning method in our proposed RHGA algorithm can avoid these issues. On the one hand, instead of using deep neural networks, our reinforcement learning method uses the traditional Q-learning algorithm~\cite{Sutton1998} to train a table. Therefore, our algorithm can solve very large instances, as we tested RHGA on instances with at most 85,900 cities. On the other hand, we combine reinforcement learning with the core search steps of LKH and EAX-GA in a reasonable way, which prevents the reduction of the efficiency. The experimental results show that RHGA significantly outperforms the newest versions of LKH and EAX-GA within similar calculation time.

\subsection{Edge Assembly Crossover Genetic Algorithm}
\label{sec_EAX}
The EAX-GA algorithm~\cite{Nagata2013} generates offspring solutions by combining edges from the two parent solutions and adding relatively few new short edges determined by a simple search procedure that is similar to the 2-opt local search. The core of EAX-GA is its edge assembly crossover (EAX) operation. Let $p_A$ and $p_B$ be two parent solutions, EAX-GA uses the EAX operation to generate $N_{ch}$ (30 by default) offsprings of $p_A$ and $p_B$, and replaces $p_A$ with the best individual among the $N_{ch}$ offsprings and $p_A$ according to an evaluation function based on the edge entropy measure~\cite{Maekawa1996}. Applying the edge entropy measure rather than the straightforward tour length measure can significantly improve the diversity of the population. Let $E_A \subset E$ and $E_B \subset E$ be the sets of edges corresponding to $p_A$ and $p_B$, the EAX operation generates offsprings through the following six steps.

\begin{itemize}
\item \textbf{Step 1:} Construct an undirected multigraph $G_{AB}=(V,E_A\cup{E_B})$ by combining all the edges of $E_A$ and $E_B$. The edges belonging to either $E_A$ or $E_B$ in $G_{AB}$ are labeled. 
\item \textbf{Step 2:} Randomly partition all edges of $G_{AB}$ into \textit{AB-cycles}, where an \textit{AB-cycle} consists of alternately linked edges of $E_A$ and $E_B$. 
\item \textbf{Step 3:} Construct an \textit{E-set} by selecting \textit{AB-cycles} according to a given selection strategy, where an \textit{E-set} is defined as the union of \textit{AB-cycles}. 
\item \textbf{Step 4:} Generate an intermediate solution from $p_A$ by removing the edges of $E_A$ and adding the edges of $E_B$ in the \textit{E-set}. Let $E_C=(E_A\backslash($\textit{E-set} $\cap$ ${E_A}))\cup{(}$\textit{E-set} $\cap$ ${E_B})$ be the set of edges in the intermediate solution. An intermediate solution consists of one or more sub-tours and may not be a feasible solution for TSP.
\item \textbf{Step 5:} Connect all sub-tours into a tour to generate a valid offspring. This step merges the smallest sub-tour (the sub-tour with the least number of edges) with other sub-tours each time. Let $U$ be the set of edges in the smallest sub-tour, the goal is to find 4-tuples of edges $\{e^*,e'^*,e''^*,e'''^*\}=\mathop{\arg\min}_{e\in{U},e'\in{E_C\backslash{U}}}\{-d(e)-d(e')+d(e'')+d(e''')\}$, where $e$ and $e'$ denote two edges to be removed, and $e''$ and $e'''$ denote two edges to be added to connect the breakpoints. Then the sub-tours are connected by $E_C\leftarrow(E_C\backslash\{e^*,e'^*\})\cup{\{e''^*,e'''^*\}}$. In particular, EAX-GA restricts the search to promising pairs of $e$ and $e'$ to reduce the search scope and improve the efficiency. For each $e\in{U}$, the candidates of $e'$ are restricted to a set of edges that satisfy the following condition: at least one end of $e'$ is among the $N_{near}$ (10 by default) closest to either end of $e$. 
\item \textbf{Step 6:} Loop steps 3-5 until $N_{ch}$ offsprings are generated. Then terminate the procedure.
\end{itemize}

Note that the metric for determining the candidates of $e'$ in Step 5 is the distance. This metric is very important since it determines the new edges that can be added to the population. In the proposed RHGA algorithm, we replace the distance metric used here with the Q-value learned by the Q-learning algorithm to improve the performance.

The EAX-GA algorithm consists of two stages. It terminates stage \uppercase\expandafter{\romannumeral1} when no improvement in the best solution is found over a period of generations, and then switches to stage \uppercase\expandafter{\romannumeral2}. Specifically, let $Gen$ be the number of generations at which no improvement in the best solution is found over the recent $1500/N_{ch}$ generations. If the value of $Gen$ has already been determined and the best solution does not improve over the last $G_{max} = Gen/10$ generations, EAX-GA terminates stage \uppercase\expandafter{\romannumeral1} and proceeds to stage \uppercase\expandafter{\romannumeral2}. Stage \uppercase\expandafter{\romannumeral2} is also terminated by the same condition (both $Gen$ and $G_{max}$ should be recalculated in this stage). 

The only difference between the two stages is the selection strategy of the \textit{E-set} (Step 3) during the EAX crossover process. In stage \uppercase\expandafter{\romannumeral1}, a single \textit{AB-cycle} is selected randomly as the \textit{E-set} without overlapping with the previous selections. Such a strategy is very simple and fast, thus can help the population converge quickly. In stage \uppercase\expandafter{\romannumeral2}, the \textit{block2} strategy~\cite{Nagata2013} is applied, which is effective in solving large TSP instances. Its basic idea is to construct an \textit{E-set} by selecting \textit{AB-cycles} so that the resulting intermediate solution consists of relatively few sub-tours and the resulting offspring consists of more edges of $p_B$. The intermediate solution with few sub-tours corresponds to an offspring that inherits its parents well, and making the offspring inherit more edges of $p_B$ can prevent the algorithm from falling into the local optima easily.


\subsection{Lin-Kernighan-Helsgaun Algorithm}
LKH uses the $k$-opt heuristic~\cite{Lin1965} as the optimization method to find high-quality solutions. The $k$-opt in LKH replaces at most $k_{max}$ (5 by default) edges in the current tour with the same number of new edges, and restricts that the edges to be added must be selected from the candidate sets, so as to reduce the search scope and improve the efficiency. This subsection introduces two important parts of LKH, i.e., the method of creating the candidate sets and the $k$-opt process.

\subsubsection{Candidate Sets in LKH}
\label{sec_alpha}
In LKH, each city has its candidate set that records several candidate cities. Let $CS^i$ be the candidate set of city $i$ ($i \in V$), LKH restricts that the edges to be added in the $k$-opt process must be selected from the set $\{(i,j) \in E | j \in CS^i \vee i \in CS^j\}$. LKH proposes an $\alpha$-value to evaluate the quality of the edges, and applies the $\alpha$-value as the metric for selecting and sorting candidate cities. The $\alpha$-value is defined from the structure of 1-tree~\cite{Held1970}. A 1-tree for the graph $G = (V, E)$ is a spanning tree on the node set $V\backslash\{v\}$ combined with two edges from $E$ incident to a node $v$ chosen arbitrarily. The minimum 1-tree is the 1-tree with the minimum length. Obviously, the length of the minimum 1-tree is a lower bound of the optimal TSP solution. The equation for calculating the $\alpha$-value of an edge $(i,j)$ is as follows:
\begin{equation}
\alpha(i,j) = L(T^+(i,j))-L(T),
\label{eq_alpha-value}
\end{equation}
where $L(T)$ is the length of the minimum 1-tree of the graph $G$, and $L(T^{+}(i,j))$ is the length of the minimum 1-tree required to contain edge $(i,j)$. The candidate set of each city in LKH records five (default value) other cities with the smallest $\alpha$-value to this city in ascending order. The advantage of the candidate set is further enhanced by adding \textit{penalties} to the cities. Details about the \textit{penalties} are referred to~\cite{Helsgaun2000}.

\begin{algorithm}[t]
\caption{$k$-opt($x_{in},\mathbf{p}_1,\mathbf{p},k,k_{max}$)}
\label{alg_kopt}
\LinesNumbered
\KwIn{input solution: $x_{in}$, starting city: $\mathbf{p}_1$, sequence of the corresponding cities: $\mathbf{p}$, current search depth: $k$, the maximum search depth: $k_{max}$}
\KwOut{output solution $x_{out}$, sequence of the corresponding cities $\mathbf{p}$}
\For {$i \leftarrow 1 : 2$}{
$\mathbf{p}_{2k} \leftarrow \mathbf{p}_{2k-1}^{i}$\;
\If{$\mathbf{p}_{2k}$ does not satisfy the constraint C-\uppercase\expandafter{\romannumeral1}}{\textbf{continue\;}}
\If{$k \geq 2 \wedge \sum_{j=1}^{k}{d(\mathbf{p}_{2j-1},\mathbf{p}_{2j})} < \sum_{j=1}^{k-1}{d(\mathbf{p}_{2j},\mathbf{p}_{2j+1})} + d(\mathbf{p}_{2k},\mathbf{p}_1)$}{
$x_{out} \leftarrow x_{in}$\;
\For{$j \leftarrow 1 : k$}{remove edge $(\mathbf{p}_{2j-1},\mathbf{p}_{2j})$ from $x_{out}$\;}
\For{$j \leftarrow 1 : k-1$}{add edge $(\mathbf{p}_{2j},\mathbf{p}_{2j+1})$ into $x_{out}$\;}
Add edge $(\mathbf{p}_{2k},\mathbf{p}_1)$ into $x_{out}$\;
\textbf{return} $(x_{out},\mathbf{p})$\;
}
\lIf{$k = k_{max}$}{\textbf{return} $(x_{in},\emptyset)$}
\For {$j \leftarrow 1 : 5$}{
$\mathbf{p}_{2k+1} \leftarrow$ the $j$-th city in $CS^{\mathbf{p}_{2k}}$\;
\If{$\mathbf{p}_{2k+1}$ doesn't satisfy constraint C-\uppercase\expandafter{\romannumeral2}}{\textbf{continue\;}}
$(x_{temp},\mathbf{p'}) \leftarrow k$-opt($x_{in},\mathbf{p}_1,\mathbf{p} \cup \{\mathbf{p}_{2k},\mathbf{p}_{2k+1}\},k+1,k_{max}$)\;
\If{$l(x_{temp}) < l(x_{in})$}{\textbf{return} $(x_{temp},\mathbf{p'})$\;}
}
}
\textbf{return} $(x_{in},\emptyset)$\;
\end{algorithm}

\subsubsection{$k$-opt in LKH}
The $k$-opt process is actually a partial depth-first search process, that the maximum depth of the search tree is restricted to $k_{max}$. The $k$-opt process starts from a starting city $\mathbf{p}_1$ (i.e., root of the search tree), then alternatively selects an edge to be removed, i.e., edge $(\mathbf{p}_{2k-1},\mathbf{p}_{2k})$, and an edge to be added, i.e., edge $(\mathbf{p}_{2k},\mathbf{p}_{2k+1})$, until the maximum search depth is reached or a $k$-opt move that can improve the current tour is found. Note that these edges are connected, thus selecting the involved edges in $k$-opt can be regarded as selecting a sequence (cycle) of cities. The selection of the cities $\mathbf{p}_{2k}$ and $\mathbf{p}_{2k+1}$ should satisfy the following constraints:

\begin{itemize}
\item \textbf{C-\uppercase\expandafter{\romannumeral1}:} for $k\geq2$, connecting $\mathbf{p}_{2k}$ back to $\mathbf{p}_1$ should result in a feasible TSP tour.
\item \textbf{C-\uppercase\expandafter{\romannumeral2}:} $\mathbf{p}_{2k+1}$ is always chosen so that $\sum\nolimits_{j=1}^{i}(d(\mathbf{p}_{2k-1},\mathbf{p}_{2k})-d(\mathbf{p}_{2k},\mathbf{p}_{2k+1}))>0$.
\end{itemize}

Let $t^1$ be a city randomly picked from the two cities connected with city $t$ in the current TSP tour, $t^2$ be the other, $l(x)$ be the length of solution $x$. The procedure of the $k$-opt process is presented in Algorithm \ref{alg_kopt}. As shown in Algorithm \ref{alg_kopt}, the $k$-opt process tries to improve the current solution by traversing the partial depth-first search tree from the root $\mathbf{p}_1$. When selecting the edge to be removed, i.e., edge $(\mathbf{p}_{2k-1},\mathbf{p}_{2k})$ (the same as selecting $\mathbf{p}_{2k}$ from $\mathbf{p}_{2k-1}$), the algorithm traverses the two cities connected with city $\mathbf{p}_{2k-1}$ in the current TSP tour (lines 1-2). When selecting the edge to be added, i.e., edge $(\mathbf{p}_{2k},\mathbf{p}_{2k+1})$ (the same as selecting $\mathbf{p}_{2k+1}$ from $\mathbf{p}_{2k}$), the algorithm traverses the candidate set of city $\mathbf{p}_{2k}$ (lines 14-15), and the constraint C-\uppercase\expandafter{\romannumeral2} is applied as a smart pruning strategy to improve the efficiency (lines 16-17). Once a $k$-opt move that can improve the current solution is found, the algorithm performs this move on $x_{in}$ and outputs the resulting solution $x_{out}$ (lines 5-12). 

\section{The Proposed Algorithm}
\label{Sec_RHGA}
In the proposed reinforced hybrid genetic algorithm (RHGA), we design a novel hybrid mechanism with a special individual as the core to combine EAX-GA with the LKH local search. The EAX-GA and LKH can boost each other with the help of the special individual. The reinforcement learning technique~\cite{Zheng2021} is combined with the key steps of both LKH and EAX-GA to further improve the hybrid genetic algorithm, by replacing the evaluation metrics for the edges used in LKH ($\alpha$-value) and EAX-GA (distance) with the learned adaptive Q-value. 

This section first introduces how Q-value (i.e., reinforcement learning) is used in RHGA, then introduces the reinforced LKH local search method (Q-LKH) in RHGA, and describes the main process of RHGA that contains the description of the proposed hybrid mechanism, and finally concludes the advantages of RHGA.


\subsection{Q-value in RHGA}
The Q-value in RHGA actually determines the candidate edges in both LKH and EAX-GA. Note that the larger the Q-value of an edge, the higher-quality of the edge. The candidate set of each city in RHGA records $K$ (25 by default) other cities with the largest Q-values to this city in descending order. When selecting an edge to be added $(\mathbf{p}_{2k},\mathbf{p}_{2k+1})$ during the $k$-opt process in the Q-LKH local search component of RHGA, $\mathbf{p}_{2k+1}$ can only be selected among the top five (default value) cities in the candidate set of $\mathbf{p}_{2k}$. Similarly, when merging two sub-tours during the offspring generating process in the EAX-GA component of RHGA, the two edges to be removed, $e$ and $e'$, must satisfy that at least one end of $e'$ is among the top ten (default value) cities in the candidate set of either end of $e$.

RHGA designs an initial Q-value for each edge to generate the initial candidate sets. Before calculating the initial Q-value, the algorithm needs to calculate the lower bound of the optimal TSP solution $L(T)$ (see Eq. \ref{eq_alpha-value}) and the $\alpha$-values corresponding to $L(T)$, by the method in LKH (see Section \ref{sec_alpha}). Then the initial Q-value for edge $(i,j)$ can be calculated by:
\begin{equation}
Q(i,j)=\frac{L(T)}{\alpha(i,j)+d(i,j)}.
\label{eq_InitQ}
\end{equation}

The initial Q-value combines the metrics of evaluating the quality of edges in both EAX-GA and LKH, i.e., the distance and $\alpha$-value. The $L(T)$ is applied to adaptively adjust the magnitude of the initial Q-value for different instances. 

The Q-value can be updated by the Q-learning algorithm during the Q-LKH component of RHGA (see details in the next subsection). Note that the EAX-GA component only uses the Q-value but does not update it. After each Q-LKH process, the candidate set of each city in RHGA will be sorted according to the updated Q-value. Therefore, the order or the elements of the top five or top ten cities in the candidate set of each city might be changed by our reinforcement learning method. In this way, our reinforcement learning method can provide better candidate edges for both the EAX-GA and LKH components of RHGA and help the algorithm learn to select appropriate edges to be added during the $k$-opt process and the sub-tour merging process.



\subsection{The Q-LKH Local Search Algorithm}
We apply the method proposed by Zheng et al.~\cite{Zheng2021} to combine Q-learning~\cite{Sutton1998} with LKH to learn the Q-value. The reinforced LKH algorithm (by Q-learning) is denoted as Q-LKH.

\begin{algorithm}[t]
\caption{Q-LKH($x_{in},k_{max},\lambda,\gamma$)}
\label{alg_QLKH}
\LinesNumbered
\KwIn{input solution $x_{in}$, the maximum search depth: $k_{max}$, learning rate: $\lambda$, reward discount factor: $\gamma$}
\KwOut{output solution: $x_{out}$}
Initialize the set of the cities that have not been selected as the starting city of the $k$-opt: $A \leftarrow \{1,2,...,n\}$, $x_{out} \leftarrow x_{in}$\;
\While{TRUE}{
\lIf{$A = \emptyset$}{\textbf{break}}
$\mathbf{p}_1 \leftarrow$ a random city in $A$, $A \leftarrow A \backslash\{\mathbf{p}_1\}$\;
$(x_{temp},\mathbf{p}) \leftarrow k$-opt($x_{out},\mathbf{p}_1,\{\mathbf{p}_1\},1,k_{max}$)\;
Update the Q-value of each state-action pair in $\mathbf{p}$ according to Eq. \ref{eq_QLearning}\;
\If{$l(x_{temp}) < l(x_{out})$}{
$x_{out} \leftarrow x_{temp}$\;
\lFor{$j \leftarrow 1 : |\mathbf{p}|$}{$A \leftarrow A \cup \{\mathbf{p}_j\}$}
}
}
Sort the candidate sets of each city in descending order of the Q-value\;
\textbf{return} $x_{out}$\;
\end{algorithm}

In Q-LKH, the reinforcement learning is combined with the core $k$-opt search process. A $k$-opt process corresponds to an episode in reinforcement learning, where the states and actions are the two endpoints of the selected edges to be added during the $k$-opt process. Specifically, for an episode $(x',\mathbf{p}) \leftarrow k$-opt($x,\mathbf{p}_1,\{\mathbf{p}_1\},1,k_{max}$), the states are the cities that are going to select the edges to be added from their candidate sets, i.e., cities $\mathbf{p}_{2k}, k \in \{1,2,...,\frac{|\mathbf{p}|}{2}-1\}$, and the actions correspond to the selection of the candidate cities, i.e., cities $\mathbf{p}_{2k+1}, k \in \{1,2,...,\frac{|\mathbf{p}|}{2}-1\}$. The reward of the state-action pair $(\mathbf{p}_{2k},\mathbf{p}_{2k+1})$ is defined as $r_k = d(\mathbf{p}_{2k-1},\mathbf{p}_{2k}) - d(\mathbf{p}_{2k},\mathbf{p}_{2k+1})$, since the $k$-opt move replaces edge $(\mathbf{p}_{2k-1},\mathbf{p}_{2k})$ with edge $(\mathbf{p}_{2k},\mathbf{p}_{2k+1})$.

The Q-LKH applies the Q-learning algorithm to update the Q-value of each state-action pair in each episode ($k$-opt process). For an episode $(x',\mathbf{p}) \leftarrow k$-opt($x,\mathbf{p}_1,\{\mathbf{p}_1\},1,k_{max}$), the Q-value of each state-action pair $(\mathbf{p}_{2k},\mathbf{p}_{2k+1})$ is updated as follows:
\vspace{-0.5em}
\begin{equation}
\begin{aligned}
Q(\mathbf{p}_{2k},\mathbf{p}_{2k+1})=(1-\lambda)&\cdot{Q(\mathbf{p}_{2k},\mathbf{p}_{2k+1}))}+\\
\lambda&\cdot[r_k+\gamma{\max\limits_{a' \in CS^{\mathbf{p}_{2k+2}}}Q(\mathbf{p}_{2k+2},a')}],
\label{eq_QLearning}
\end{aligned}
\end{equation}
where $\lambda$ is the learning rate, and $\gamma$ is the reward discount factor.

The procedure of the Q-LKH local search is presented in Algorithm \ref{alg_QLKH}. Q-LKH algorithm uses the $k$-opt heuristic (Algorithm \ref{alg_kopt}) to improve the current solution $x_{out}$ until the local optimum is reached (lines 2-9), i.e., $x_{out}$ cannot be improved by the $k$-opt heuristic starting from any starting city $\mathbf{p}_1 \in \{1,2,...,n\}$ (line 3). Once the current solution is improved by a $k$-opt move (lines 7-8), each involved city can be selected as the root $\mathbf{p}_1$ again (line 9). Q-LKH updates the Q-value after each $k$-opt process (line 6), and reorders the candidate sets of each city at the end of the algorithm (line 10).


\begin{algorithm}[t]
\caption{RHGA($N_{pop},N_{ch},k_{max},\lambda,\gamma,M_{gen},\textit{OPT}$)}
\label{alg_RHGA}
\LinesNumbered 
\KwIn{population size: $N_{pop}$, number of offsprings produced by a pair of parents: $N_{ch}$, the maximum search depth: $k_{max}$, learning rate: $\lambda$, reward decay factor: $\gamma$, number of generations to perform Case 3: $M_{gen}$, length of the optimal solution: \textit{OPT}}
\KwOut{output solution: $x_{out}$}
Generate the initial candidate sets according to the initial Q-value (Eq. \ref{eq_InitQ})\;
$\{x_1,x_2,...,x_{N_{pop}}\}\leftarrow$ \textit{Generate\_Initial\_Pop}()\;
Initialize $l^1_{old}\leftarrow+\infty$, $l^{best}_{old}\leftarrow+\infty$, $num\leftarrow0$\;
\While{a termination condition is not satisfied}{
$x_{best}\leftarrow\mathop{\arg\min}_{x_i \in \{x_2,...,x_{N_{pop}}\}}l(x_i)$\;
\lIf {$l(x_1) =$ \textit{OPT} $\vee$ $l(x_{best}) = $ \textit{OPT}}{\textbf{break}}
$num\leftarrow num+1$\;
\If {$l(x_1)<l^1_{old}$}{
$x_1\leftarrow$ Q-LKH($x_1,k_{max}\lambda,\gamma$)\;
$l^1_{old}\leftarrow l(x_1)$, $num\leftarrow0$\;
}
\If {$l(x_{best})<l(x_1) \wedge l(x_{best})<l^{best}_{old}$}{
$l^{best}_{old}\leftarrow l(x_{best})$\;
$x_{temp}\leftarrow$ Q-LKH($x_{best},k_{max},\lambda,\gamma$)\;
\If {$l(x_{temp})<l(x_{best})$}{
$x_1\leftarrow x_{temp}$, $l^1_{old}\leftarrow l(x_1)$, $num\leftarrow0$\;
}
}
\If {$num\geq{M_{gen}}$}{
$x_r\leftarrow$ a random individual in $\{x_2,...,x_{N_{pop}}\}$\;
$x_{temp}\leftarrow$ Q-LKH($x_r,k_{max},\lambda,\gamma$), $num\leftarrow0$\;
\If {$l(x_{temp})<l(x_1)$}{
$x_1\leftarrow x_{temp}$, $l^1_{old}\leftarrow l(x_1)$\;
}
}
$rp(\cdot)\leftarrow$ a random permutation of $\{1,2,...,N_{pop}\}$\;
\For {$i\leftarrow 1 : N_{pop}$}{
$p_A\leftarrow x_{rp(i)}$, $p_B\leftarrow x_{rp(i+1)}$\; $\{c_1,c_2,...,c_{N_{ch}}\}\leftarrow$ \textit{EAX}$(p_A,p_B)$\;
$x_{rp(i)}\leftarrow$ \textit{Select\_Survive}$(c_1,...,c_{N_{ch}},p_A)$\;
}
}
\lIf{$l(x_1)<l(x_{best})$}{\textbf{return} $x_1$}
\lElse{\textbf{return} $x_{best}$}
\end{algorithm}

\subsection{Main Process of RHGA}
\label{sec_main_RHGA}

The main flow of RHGA is presented in Algorithm \ref{alg_RHGA}. In the initialization phase of RHGA (lines 1-3), the initial candidate set of each city is generated according to the initial Q-value calculated by Eq. \ref{eq_InitQ}, and the initial population with $N_{pop}$ (300 by default) individuals is generated by the \textit{Generate\_Initial\_Pop}() function, which is a greedy 2-opt local search method used in EAX-GA~\cite{Nagata2013}. Note that the candidate sets and the Q-value are regarded as the global information in the entire RHGA algorithm. 

In the improvement phase of RHGA (lines 4-25), the Q-LKH local search algorithm and the EAX genetic algorithm are used to improve the population alternatively. In order to prevent the reduction of population diversity and algorithm efficiency, there is only one special individual (i.e., $x_1$) that can be improved by the Q-LKH local search algorithm. Specifically, before the procedure of the genetic algorithm (lines 21-25) at each generation, RHGA tries to improve the special individual $x_1$ by the Q-LKH local search algorithm in the following three cases: 


\begin{itemize}
\item \textbf{Case 1}: (lines 8-10) When $x_1$ is just initialized or $x_1$ was improved by EAX-GA at the last generation, i.e., when $x_1$ may not be a local optimal solution for the Q-LKH local search algorithm. In this case, the Q-LKH algorithm will try to improve the special individual $x_1$. 

\item \textbf{Case 2}: (lines 11-15) When the tour length of the best individual $x_{best}$ in the population other than $x_1$ is shorter than that of $x_1$, and $x_{best}$ has not been calculated by Q-LKH. In this case, the Q-LKH algorithm will try to improve $x_{best}$. If $x_{best}$ can be improved, replace $x_1$ with the improved solution, and $x_{best}$ will not change. 

\item \textbf{Case 3}: (lines 16-20) When $x_1$ has not been improved for $M_{gen}$ generations. Note that the counter \textit{num} will always be initialized to zero (no matter whether $x_1$ can be improved). In this case, RHGA randomly selects an individual $x_r$ in the population ($x_r\neq{x_1}$). If Q-LKH can improve $x_r$ and the improved tour is better than $x_1$, the improved tour will replace $x_1$, and $x_r$ will not change. 
\end{itemize}

The design of applying the Q-LKH to improve the special individual in the above three cases is reasonable and effective. Firstly, in the first two cases, the Q-LKH is prohibited from performing on its local optimal solutions to improve the efficiency, since Q-LKH can hardly improve the local optimal solution calculated by itself. Secondly, in Case 2, the individual $x_{best}$ with a shorter length than $x_1$ is a very high-quality initial solution for the Q-LKH. Because $x_{best}$ is better than the local optimal solution of $x_1$, and it may not be a local optimum for Q-LKH. Thus performing Q-LKH on $x_{best}$ in Case 2 is necessary, and may obtain the near-optimal or even the optimal solution. Thirdly, in Case 3, various individuals can provide high-quality and diverse initial solutions for Q-LKH to escape from the local optima.

After the local search process in each generation we have the EAX genetic process (lines 21-25). During this process, Each individual in the population is selected once as parent $p_A$ and once as parent $p_B$, in a random order (lines 21-23). The algorithm applies the methods described in Section \ref{sec_EAX} to use the EAX crossover operation represented by  function \textit{EAX}() to parents $p_A$ and $p_B$ to produce $N_{ch}$ offsprings (line 24), and then selects the surviving individual among the offsprings and $p_A$ (line 25).


The RHGA algorithm also consists of two stages as EAX-GA does. The termination conditions of the two stages in RHGA are the same as those in EAX-GA (see Section \ref{sec_EAX}). Moreover, if the definite optimal solution of the TSP instance is known, the input parameter \textit{OPT} is set to the length of the optimal solution, otherwise zero. RHGA also terminates when the definite optimal solution is found (line 6).

\subsection{Advantages of RHGA}
This subsection illustrates the advantages of the proposed RHGA algorithm, i.e., why the RHGA is effective and better than the baseline algorithms (EAX-GA and LKH)? The advantages of RHGA over the baselines include the mechanism of the hybrid genetic algorithm and the impact of reinforcement learning.

\subsubsection{Mechanism of the Hybrid Genetic Algorithm}
The combination of EAX-GA and LKH by the proposed mechanism can boost the performance of each other. 
For the EAX-GA, the special individual $x_1$ can spread good genes (the candidate edges in LKH) to the population, and lead the population to converge to better solutions. For the LKH, the population can help the special individual $x_1$ escape from the local optima of LKH, and provide higher-quality and more diverse initial tours than the initial tours generated by the heuristic in LKH~\cite{Helsgaun2000,Helsgaun2009}. The hybrid mechanism in RHGA can improve the baseline algorithms without reducing the population diversity and algorithm efficiency, since there is only one special individual in the population. The experimental results also demonstrate that setting only one special individual is reasonable and efficient.

Moreover, the combination of EAX-GA and LKH can combine their advantages and overcome their disadvantages (their pros and cons are described in Section \ref{sec_intro}). That is, RHGA can solve the TSP instances with tens to hundreds of cities as well as or better than EAX-GA does, and can obtain solutions of acceptable quality within reasonable calculation time when solving the TSP instances with various scales like LKH does.

\subsubsection{Impact of the Reinforcement Learning}
As indicated by the results in~\cite{Helsgaun2000}, the $\alpha$-value outperforms the distance in determining the candidate cities or evaluating the quality of the edges. As indicated by the results in~\cite{Zheng2021}, the Q-value is a better choice than the $\alpha$-value. So why not replace the $\alpha$-value metric used in LKH and the distance metric used in the sub-tours merging process in EAX-GA with our learned adaptive Q-value?

In the RHGA algorithm, the reinforcement learning is incorporated into both the local search process and the population optimization process in RHGA, by learning an adaptive Q-value to select and sort the candidate edges in both LKH and EAX-GA. Note that the initial candidate edges determined by the initial Q-value (Eq. \ref{eq_InitQ}) are better than the candidate edges determined by distance metric or $\alpha$-value (see experimental results in Section \ref{Sec_Exp}). The reinforcement learning can further improve the quality of the candidate edges by updating the Q-value and adjusting the candidate sets. In particular, the experimental results demonstrate that the order of the performance of EAX-GA with different metrics with 
decaying quality is: adaptive Q-value (updated by Eq. \ref{eq_QLearning}), initial Q-value (Eq. \ref{eq_InitQ}), $\alpha$-value, and finally the distance.

\section{Experimental Results}
\label{Sec_Exp}
This section presents the computational results and comparisons of RHGA, EAX-GA, LKH, and NeuroLKH~\cite{Xin2021}. The results show that RHGA significantly outperforms the other three algorithms. We first introduce the experimental setup, the benchmark instances and the baseline algorithms, then present the experimental results.

\subsection{Experimental Setup}
The experiments of RHGA were implemented in C++ and compiled by g++ with -O3 option. All the algorithms in the experiments were run on a server using an Intel® Xeon® E5-2650 v3 2.30 GHz 10-core CPU and 256 GB RAM, running Ubuntu 16.04 Linux operation system. The algorithms were all run on a single core. The parameters related to genetic algorithm in RHGA are set to be the same as the default settings in EAX-GA~\cite{Nagata2013}, i.e., $N_{pop}=300$, $N_{ch}=30$. Other parameters are set as follows: $\lambda=0.1$, $\gamma=0.9$, $M_{gen}=10(\log_{10}{n}-1)$ (i.e., $M_{gen}=20/30/40$ when $n=10^3/10^4/10^5$). To reduce the variance in the results, we run each algorithm in the experiments 10 times on each TSP instance.

\subsection{Benchmark Instances}
The RHGA algorithm was tested on all the TSP instances with the number of cities ranging from 1,000 to 85,900 cities, with a total of 138, in the well-known and widely used benchmark sets for the TSP: TSPLIB\footnote{http://comopt.ifi.uni-heidelberg.de/software/TSPLIB95}, National TSP benchmarks\footnote{http://www.math.uwaterloo.ca/tsp/world/countries.html}, and VLSI TSP benchmarks\footnote{http://www.math.uwaterloo.ca/tsp/vlsi/index.html}. Note that the number in each instance's name indicates the number of cities in that instance. 

In order to make a clear comparison, we divide the 138 instances into \textit{small} and \textit{large} according to the instance scale. That is, an instance with less than 20,000 cities is considered to be \textit{small}, otherwise \textit{large}. There are a total of 111 \textit{small} instances and 27 \textit{large} instances among all the 138 tested instances. 

Moreover, we further divide the 138 instances into the following three categories according to their difficulty:

\begin{itemize}
\item \textit{Easy}: An instance is \textit{easy} when both RHGA and EAX-GA (with the default settings) can obtain the best-known solution of this instance in each of the 10 runs (i.e., the worst solutions of RHGA and EAX-GA in 10 runs are all equal to the best-known solution when solving this instance). There are a total of 60 \textit{easy} instances among all the 138 tested instances.
\item \textit{Medium}: An instance is \textit{medium} when it satisfies the following two conditions: 1) the best solutions of RHGA and EAX-GA in 10 runs are all equal to the best-known solution of this instance. 2) At least one of the worst solutions of RHGA and EAX-GA is not equal to the best-known solution of this instance. There are a total of 62 \textit{medium} instances among all the 138 tested instances.
\item \textit{Hard}: An instance is \textit{hard} if at least one of the best solutions of RHGA and EAX-GA is not equal to the best-known solution of this instance. There are a total of 16 \textit{hard} instances among all the 138 tested instances.
\end{itemize}


\begin{table*}[t]
\centering
\caption{Comparison of RHGA, NeuroLKH\_R, and NeuroLKH\_M. Best results appear in bold.}
\label{table_NeuroLKH}
\scalebox{0.65}{\begin{tabular}{lrrrrrrrrrrrr} \toprule
\multirow{2}{*}{Instance} & \multirow{2}{*}{BKS} & \multicolumn{3}{c}{RHGA}                                       &  & \multicolumn{3}{c}{NeuroLKH\_R}                             &  & \multicolumn{3}{c}{NeuroLKH\_M}                              \\ \cline{3-5} \cline{7-9} \cline{11-13} 
                          &                      & Best (gap\%)              & Average (gap\%)             & Time   &  & Best (gap\%)             & Average (gap\%)           & Time   &  & Best (gap\%)             & Average (gap\%)            & Time   \\ \hline
u1060                     & 224094               & \textbf{224094 (0.0000)}  & \textbf{224094.0 (0.0000)}  & 28.4   &  & \textbf{224094 (0.0000)} & 224099.1 (0.0023)         & 26.8   &  & \textbf{224094 (0.0000)} & \textbf{224094.0 (0.0000)} & 21.3   \\
vm1084                    & 239297               & \textbf{239297 (0.0000)}  & \textbf{239297.0 (0.0000)}  & 19.1   &  & \textbf{239297 (0.0000)} & 239379.5 (0.0345)         & 16.5   &  & \textbf{239297 (0.0000)} & 239326.4 (0.0123)          & 27.9   \\
pcb1173                   & 56892                & \textbf{56892 (0.0000)}   & \textbf{56892.0 (0.0000)}   & 21.8   &  & \textbf{56892 (0.0000)}  & 56892.5 (0.0009)          & 9.2    &  & \textbf{56892 (0.0000)}  & 56893.0 (0.0018)           & 7.7    \\
d1291                     & 50801                & \textbf{50801 (0.0000)}   & \textbf{50801.0 (0.0000)}   & 10.6   &  & \textbf{50801 (0.0000)}  & 50803.4 (0.0047)          & 11.4   &  & \textbf{50801 (0.0000)}  & 50808.2 (0.0142)           & 6.4    \\
rl1304                    & 252948               & \textbf{252948 (0.0000)}  & \textbf{252948.0 (0.0000)}  & 9.3    &  & \textbf{252948 (0.0000)} & 252953.1 (0.0020)         & 9.2    &  & \textbf{252948 (0.0000)} & 252958.2 (0.0040)          & 20.9   \\
rl1323                    & 270199               & \textbf{270199 (0.0000)}  & \textbf{270199.0 (0.0000)}  & 12.7   &  & \textbf{270199 (0.0000)} & 270247.9 (0.0181)         & 16.8   &  & \textbf{270199 (0.0000)} & 270204.4 (0.0020)          & 24.2   \\
nrw1379                   & 56638                & \textbf{56638 (0.0000)}   & \textbf{56638.0 (0.0000)}   & 59.9   &  & \textbf{56638 (0.0000)}  & 56638.5 (0.0009)          & 20.8   &  & \textbf{56638 (0.0000)}  & \textbf{56638.0 (0.0000)}  & 22.9   \\
dca1389                   & 5085                 & \textbf{5085 (0.0000)}    & \textbf{5085.0 (0.0000)}    & 31.6   &  & 5087 (0.0393)            & 5087.0 (0.0393)           & 19.6   &  & \textbf{5085 (0.0000)}   & 5086.5 (0.0295)            & 11.5   \\
fl1400                    & 20127                & \textbf{20127 (0.0000)}   & \textbf{20127.0 (0.0000)}   & 74.2   &  & 20185 (0.2882)           & 20185.0 (0.2882)          & 692.5  &  & 20189 (0.3080)           & 20189.0 (0.3080)           & 564.2  \\
dja1436                   & 5257                 & \textbf{5257 (0.0000)}    & \textbf{5257.0 (0.0000)}    & 25.2   &  & \textbf{5257 (0.0000)}   & 5257.2 (0.0038)           & 41.0   &  & \textbf{5257 (0.0000)}   & \textbf{5257.0 (0.0000)}   & 67.6   \\
fra1488                   & 4264                 & \textbf{4264 (0.0000)}    & \textbf{4264.0 (0.0000)}    & 18.1   &  & \textbf{4264 (0.0000)}   & 4264.1 (0.0023)           & 34.1   &  & \textbf{4264 (0.0000)}   & \textbf{4264.0 (0.0000)}   & 2.0    \\
fl1577                    & 22249                & \textbf{22249 (0.0000)}   & \textbf{22249.0 (0.0000)}   & 67.6   &  & 22256 (0.0315)           & 22256.0 (0.0315)          & 152.8  &  & 22698 (2.0181)           & 22698.0 (2.0181)           & 613.8  \\
rbv1583                   & 5387                 & \textbf{5387 (0.0000)}    & \textbf{5387.0 (0.0000)}    & 41.8   &  & \textbf{5387 (0.0000)}   & \textbf{5387.0 (0.0000)}  & 36.2   &  & \textbf{5387 (0.0000)}   & 5387.1 (0.0019)            & 33.1   \\
fnb1615                   & 4956                 & \textbf{4956 (0.0000)}    & 4956.1 (0.0020)             & 48.4   &  & \textbf{4956 (0.0000)}   & 4957.5 (0.0303)           & 114.2  &  & \textbf{4956 (0.0000)}   & \textbf{4956.0 (0.0000)}   & 36.0   \\
rw1621                    & 26051                & \textbf{26051 (0.0000)}   & \textbf{26051.0 (0.0000)}   & 47.8   &  & 26056 (0.0192)           & 26056.0 (0.0192)          & 735.7  &  & 26077 (0.0998)           & 26077.0 (0.0998)           & 452.9  \\
d1655                     & 62128                & \textbf{62128 (0.0000)}   & \textbf{62128.0 (0.0000)}   & 47.3   &  & \textbf{62128 (0.0000)}  & 62128.2 (0.0003)          & 44.0   &  & \textbf{62128 (0.0000)}  & \textbf{62128.0 (0.0000)}  & 25.1   \\
vm1748                    & 336556               & \textbf{336556 (0.0000)}  & \textbf{336556.0 (0.0000)}  & 53.3   &  & \textbf{336556 (0.0000)} & 336628.0 (0.0214)         & 42.4   &  & \textbf{336556 (0.0000)} & \textbf{336556.0 (0.0000)} & 36.7   \\
djc1785                   & 6115                 & \textbf{6115 (0.0000)}    & \textbf{6115.0 (0.0000)}    & 61.0   &  & \textbf{6115 (0.0000)}   & 6115.5 (0.0082)           & 77.9   &  & \textbf{6115 (0.0000)}   & 6115.6 (0.0098)            & 42.0   \\
u1817                     & 57201                & \textbf{57201 (0.0000)}   & \textbf{57209.1 (0.0142)}   & 51.0   &  & \textbf{57201 (0.0000)}  & 57221.3 (0.0355)          & 159.2  &  & \textbf{57201 (0.0000)}  & 57239.3 (0.0670)           & 109.8  \\
rl1889                    & 316536               & \textbf{316536 (0.0000)}  & \textbf{316536.0 (0.0000)}  & 32.7   &  & 316638 (0.0322)          & 316646.8 (0.0350)         & 44.9   &  & 316638 (0.0322)          & 316650.0 (0.0360)          & 58.1   \\
dcc1911                   & 6396                 & \textbf{6396 (0.0000)}    & \textbf{6396.0 (0.0000)}    & 53.7   &  & \textbf{6396 (0.0000)}   & 6396.2 (0.0031)           & 115.0  &  & \textbf{6396 (0.0000)}   & 6396.8 (0.0125)            & 22.5   \\
dkd1973                   & 6421                 & \textbf{6421 (0.0000)}    & \textbf{6421.0 (0.0000)}    & 46.2   &  & \textbf{6421 (0.0000)}   & 6422.0 (0.0156)           & 241.5  &  & \textbf{6421 (0.0000)}   & \textbf{6421.0 (0.0000)}   & 46.7   \\
mu1979                    & 86891                & \textbf{86891 (0.0000)}   & \textbf{86891.0 (0.0000)}   & 160.4  &  & 87191 (0.3453)           & 87211.5 (0.3689)          & 412.2  &  & 87021 (0.1496)           & 87021.0 (0.1496)           & 722.7  \\
d2103                     & 80450                & \textbf{80450 (0.0000)}   & \textbf{80450.0 (0.0000)}   & 48.4   &  & 80454 (0.0050)           & 80454.0 (0.0050)          & 21.7   &  & 80459 (0.0112)           & 80459.0 (0.0112)           & 837.0  \\
u2152                     & 64253                & \textbf{64253 (0.0000)}   & \textbf{64253.0 (0.0000)}   & 61.4   &  & \textbf{64253 (0.0000)}  & 64264.4 (0.0177)          & 60.1   &  & \textbf{64253 (0.0000)}  & 64255.8 (0.0044)           & 244.3  \\
xqc2175                   & 6830                 & \textbf{6830 (0.0000)}    & \textbf{6830.0 (0.0000)}    & 76.6   &  & \textbf{6830 (0.0000)}   & 6830.5 (0.0073)           & 122.5  &  & 6831 (0.0146)            & 6831.0 (0.0146)            & 23.9   \\
bck2217                   & 6764                 & \textbf{6764 (0.0000)}    & \textbf{6764.3 (0.0044)}    & 81.9   &  & 6765 (0.0148)            & 6765.0 (0.0148)           & 35.6   &  & \textbf{6764 (0.0000)}   & \textbf{6764.3 (0.0044)}   & 53.2   \\
xpr2308                   & 7219                 & \textbf{7219 (0.0000)}    & \textbf{7219.1 (0.0014)}    & 80.4   &  & \textbf{7219 (0.0000)}   & 7219.5 (0.0069)           & 47.3   &  & \textbf{7219 (0.0000)}   & 7219.9 (0.0125)            & 163.1  \\
ley2323                   & 8352                 & \textbf{8352 (0.0000)}    & \textbf{8352.0 (0.0000)}    & 49.2   &  & 8355 (0.0359)            & 8358.4 (0.0766)           & 85.1   &  & 8355 (0.0359)            & 8355.0 (0.0359)            & 53.7   \\
dea2382                   & 8017                 & \textbf{8017 (0.0000)}    & \textbf{8017.0 (0.0000)}    & 71.6   &  & 8018 (0.0125)            & 8019.0 (0.0249)           & 122.4  &  & \textbf{8017 (0.0000)}   & 8017.1 (0.0012)            & 117.3  \\
pds2566                   & 7643                 & \textbf{7643 (0.0000)}    & \textbf{7643.0 (0.0000)}    & 102.6  &  & \textbf{7643 (0.0000)}   & 7643.7 (0.0092)           & 135.5  &  & \textbf{7643 (0.0000)}   & 7643.3 (0.0039)            & 66.0   \\
mlt2597                   & 8071                 & \textbf{8071 (0.0000)}    & \textbf{8071.0 (0.0000)}    & 47.6   &  & \textbf{8071 (0.0000)}   & 8071.4 (0.0050)           & 142.2  &  & \textbf{8071 (0.0000)}   & \textbf{8071.0 (0.0000)}   & 15.2   \\
bch2762                   & 8234                 & \textbf{8234 (0.0000)}    & 8234.1 (0.0012)             & 131.6  &  & \textbf{8234 (0.0000)}   & \textbf{8234.0 (0.0000)}  & 100.0  &  & \textbf{8234 (0.0000)}   & 8234.6 (0.0073)            & 62.2   \\
irw2802                   & 8423                 & \textbf{8423 (0.0000)}    & \textbf{8423.0 (0.0000)}    & 86.1   &  & \textbf{8423 (0.0000)}   & 8424.0 (0.0119)           & 96.8   &  & \textbf{8423 (0.0000)}   & \textbf{8423.0 (0.0000)}   & 114.5  \\
dbj2924                   & 10128                & \textbf{10128 (0.0000)}   & \textbf{10128.0 (0.0000)}   & 127.1  &  & \textbf{10128 (0.0000)}  & 10128.1 (0.0010)          & 138.3  &  & \textbf{10128 (0.0000)}  & 10128.7 (0.0069)           & 60.2   \\
xva2993                   & 8492                 & \textbf{8492 (0.0000)}    & \textbf{8492.0 (0.0000)}    & 128.1  &  & \textbf{8492 (0.0000)}   & \textbf{8492.0 (0.0000)}  & 66.8   &  & \textbf{8492 (0.0000)}   & 8492.4 (0.0047)            & 156.0  \\
pcb3038                   & 137694               & \textbf{137694 (0.0000)}  & \textbf{137694.0 (0.0000)}  & 151.1  &  & \textbf{137694 (0.0000)} & 137694.8 (0.0006)         & 217.6  &  & \textbf{137694 (0.0000)} & 137694.8 (0.0006)          & 202.7  \\
pia3056                   & 8258                 & \textbf{8258 (0.0000)}    & 8258.5 (0.0061)             & 156.4  &  & 8261 (0.0363)            & 8261.5 (0.0424)           & 70.2   &  & \textbf{8258 (0.0000)}   & \textbf{8258.2 (0.0024)}   & 164.5  \\
dke3097                   & 10539                & \textbf{10539 (0.0000)}   & \textbf{10539.0 (0.0000)}   & 127.4  &  & \textbf{10539 (0.0000)}  & 10539.2 (0.0019)          & 236.6  &  & \textbf{10539 (0.0000)}  & \textbf{10539.0 (0.0000)}  & 109.5  \\
lsn3119                   & 9114                 & \textbf{9114 (0.0000)}    & \textbf{9114.0 (0.0000)}    & 120.3  &  & \textbf{9114 (0.0000)}   & 9115.0 (0.0110)           & 249.4  &  & \textbf{9114 (0.0000)}   & 9114.1 (0.0011)            & 41.6   \\
lta3140                   & 9517                 & \textbf{9517 (0.0000)}    & \textbf{9517.0 (0.0000)}    & 134.5  &  & 9518 (0.0105)            & 9518.0 (0.0105)           & 84.4   &  & \textbf{9517 (0.0000)}   & 9517.2 (0.0021)            & 90.0   \\
fdp3256                   & 10008                & \textbf{10008 (0.0000)}   & \textbf{10008.1 (0.0010)}   & 127.4  &  & \textbf{10008 (0.0000)}  & 10008.5 (0.0050)          & 572.6  &  & \textbf{10008 (0.0000)}  & 10011.0 (0.0300)           & 33.6   \\
beg3293                   & 9772                 & \textbf{9772 (0.0000)}    & \textbf{9772.2 (0.0020)}    & 131.1  &  & 9773 (0.0102)            & 9774.0 (0.0205)           & 423.4  &  & \textbf{9772 (0.0000)}   & 9772.7 (0.0072)            & 121.0  \\
nu3496                    & 96132                & \textbf{96132 (0.0000)}   & \textbf{96132.1 (0.0001)}   & 169.1  &  & 96285 (0.1592)           & 96285.0 (0.1592)          & 1722.0 &  & 96167 (0.0364)           & 96167.0 (0.0364)           & 1088.5 \\
fjs3649                   & 9272                 & \textbf{9272 (0.0000)}    & \textbf{9272.0 (0.0000)}    & 176.3  &  & 9286 (0.1510)            & 9289.0 (0.1833)           & 430.0  &  & 9274 (0.0216)            & 9278.3 (0.0679)            & 182.6  \\
fjr3672                   & 9601                 & \textbf{9601 (0.0000)}    & \textbf{9601.0 (0.0000)}    & 158.2  &  & 9604 (0.0312)            & 9608.0 (0.0729)           & 470.8  &  & 9602 (0.0104)            & 9602.0 (0.0104)            & 372.4  \\
dlb3694                   & 10959                & \textbf{10959 (0.0000)}   & \textbf{10959.3 (0.0027)}   & 216.7  &  & \textbf{10959 (0.0000)}  & 10959.5 (0.0046)          & 228.0  &  & 10960 (0.0091)           & 10960.0 (0.0091)           & 34.0   \\
ltb3729                   & 11821                & \textbf{11821 (0.0000)}   & \textbf{11821.0 (0.0000)}   & 171.6  &  & 11822 (0.0085)           & 11822.5 (0.0127)          & 511.8  &  & \textbf{11821 (0.0000)}  & 11821.9 (0.0076)           & 230.5  \\
fl3795                    & 28772                & \textbf{28772 (0.0000)}   & \textbf{28777.6 (0.0195)}   & 428.3  &  & 30623 (6.4333)           & 30623.0 (6.4333)          & 1981.5 &  & 29556 (2.7249)           & 29556.0 (2.7249)           & 176.4  \\
xqe3891                   & 11995                & \textbf{11995 (0.0000)}   & \textbf{11996.1 (0.0092)}   & 228.8  &  & 11998 (0.0250)           & 11998.0 (0.0250)          & 82.5   &  & 11997 (0.0167)           & 11997.0 (0.0167)           & 138.6  \\
xua3937                   & 11239                & \textbf{11239 (0.0000)}   & \textbf{11239.0 (0.0000)}   & 134.8  &  & \textbf{11239 (0.0000)}  & \textbf{11239.0 (0.0000)} & 73.1   &  & \textbf{11239 (0.0000)}  & 11239.4 (0.0036)           & 213.6  \\
dkc3938                   & 12503                & \textbf{12503 (0.0000)}   & \textbf{12503.0 (0.0000)}   & 187.2  &  & 12506 (0.0240)           & 12506.0 (0.0240)          & 1410.1 &  & 12504 (0.0080)           & 12504.0 (0.0080)           & 123.1  \\
dkf3954                   & 12538                & \textbf{12538 (0.0000)}   & \textbf{12538.0 (0.0000)}   & 180.3  &  & \textbf{12538 (0.0000)}  & 12539.2 (0.0096)          & 190.1  &  & \textbf{12538 (0.0000)}  & \textbf{12538.0 (0.0000)}  & 106.1  \\
bgb4355                   & 12723                & \textbf{12723 (0.0000)}   & \textbf{12723.0 (0.0000)}   & 238.1  &  & 12725 (0.0157)           & 12727.5 (0.0354)          & 766.5  &  & \textbf{12723 (0.0000)}  & 12724.5 (0.0118)           & 253.6  \\
bgd4396                   & 13009                & \textbf{13009 (0.0000)}   & \textbf{13009.0 (0.0000)}   & 205.0  &  & 13011 (0.0154)           & 13013.7 (0.0361)          & 314.1  &  & \textbf{13009 (0.0000)}  & 13010.2 (0.0092)           & 182.0  \\
frv4410                   & 10711                & \textbf{10711 (0.0000)}   & \textbf{10711.0 (0.0000)}   & 224.7  &  & \textbf{10711 (0.0000)}  & 10713.5 (0.0233)          & 360.3  &  & \textbf{10711 (0.0000)}  & 10711.4 (0.0037)           & 207.3  \\
bgf4475                   & 13221                & \textbf{13221 (0.0000)}   & \textbf{13221.0 (0.0000)}   & 222.7  &  & 13230 (0.0681)           & 13230.5 (0.0719)          & 689.3  &  & \textbf{13221 (0.0000)}  & 13221.8 (0.0061)           & 540.9  \\
ca4663                    & 1290319              & \textbf{1290319 (0.0000)} & \textbf{1290319.0 (0.0000)} & 395.1  &  & 1290807 (0.0378)         & 1291931.0 (0.1249)        & 557.4  &  & 1290382 (0.0049)         & 1290597.5 (0.0216)         & 398.7  \\
xqd4966                   & 15316                & \textbf{15316 (0.0000)}   & \textbf{15316.0 (0.0000)}   & 325.7  &  & 15344 (0.1828)           & 15344.0 (0.1828)          & 2406.1 &  & 15318 (0.0131)           & 15318.0 (0.0131)           & 1079.2 \\
fqm5087                   & 13029                & \textbf{13029 (0.0000)}   & \textbf{13029.0 (0.0000)}   & 428.4  &  & 13057 (0.2149)           & 13057.0 (0.2149)          & 2416.4 &  & 13035 (0.0461)           & 13035.0 (0.0461)           & 877.7  \\
fea5557                   & 15445                & \textbf{15445 (0.0000)}   & \textbf{15445.0 (0.0000)}   & 293.8  &  & 15448 (0.0194)           & 15450.0 (0.0324)          & 637.3  &  & \textbf{15445 (0.0000)}  & 15446.0 (0.0065)           & 1280.9 \\
rl5915                    & 565530               & \textbf{565530 (0.0000)}  & \textbf{565530.0 (0.0000)}  & 289.7  &  & 566217 (0.1215)          & 566608.5 (0.1907)         & 1140.6 &  & 565585 (0.0097)          & 565585.0 (0.0097)          & 692.2  \\
rl5934                    & 556045               & \textbf{556045 (0.0000)}  & 556072.3 (0.0049)           & 476.1  &  & \textbf{556045 (0.0000)} & 556058.8 (0.0025)         & 471.8  &  & \textbf{556045 (0.0000)} & \textbf{556045.9 (0.0002)} & 345.5  \\
tz6117                    & 394718               & \textbf{394718 (0.0000)}  & 394721.2 (0.0008)           & 694.0  &  & 395193 (0.1203)          & 395193.0 (0.1203)         & 2248.3 &  & 394720 (0.0005)          & \textbf{394720.0 (0.0005)} & 1185.9 \\
xsc6880                   & 21535                & \textbf{21535 (0.0000)}   & \textbf{21535.1 (0.0005)}   & 641.7  &  & 21544 (0.0418)           & 21546.0 (0.0511)          & 492.0  &  & 21541 (0.0279)           & 21541.0 (0.0279)           & 730.6  \\
eg7146                    & 172386               & \textbf{172386 (0.0000)}  & \textbf{172386.0 (0.0000)}  & 1199.0 &  & 173594 (0.7008)          & 173611.0 (0.7106)         & 1636.9 &  & 173394 (0.5847)          & 173394.0 (0.5847)          & 2547.0 \\
bnd7168                   & 21834                & \textbf{21834 (0.0000)}   & \textbf{21834.0 (0.0000)}   & 396.6  &  & 21841 (0.0321)           & 21842.0 (0.0366)          & 704.1  &  & 21838 (0.0183)           & 21838.0 (0.0183)           & 1774.3 \\
lap7454                   & 19535                & \textbf{19535 (0.0000)}   & \textbf{19535.0 (0.0000)}   & 457.2  &  & 19544 (0.0461)           & 19545.0 (0.0512)          & 1489.6 &  & \textbf{19535 (0.0000)}  & 19535.5 (0.0026)           & 591.8  \\
ym7663                    & 238314               & \textbf{238314 (0.0000)}  & \textbf{238314.0 (0.0000)}  & 1071.9 &  & 238811 (0.2085)          & 238811.0 (0.2085)         & 3623.9 &  & 238430 (0.0487)          & 238430.0 (0.0487)          & 3655.6 \\
pm8079                    & 114855               & \textbf{114855 (0.0000)}  & \textbf{114855.3 (0.0003)}  & 1301.2 &  & 115011 (0.1358)          & 115011.0 (0.1358)         & 3858.3 &  & 115183 (0.2856)          & 115183.0 (0.2856)          & 3569.4 \\
ida8197                   & 22338                & \textbf{22338 (0.0000)}   & 22338.1 (0.0004)            & 424.8  &  & 22348 (0.0448)           & 22348.0 (0.0448)          & 531.2  &  & \textbf{22338 (0.0000)}  & \textbf{22338.0 (0.0000)}  & 211.0  \\
ei8246                    & 206171               & \textbf{206171 (0.0000)}  & \textbf{206171.6 (0.0003)}  & 1346.3 &  & 206179 (0.0039)          & 206179.0 (0.0039)         & 1844.0 &  & \textbf{206171 (0.0000)} & 206173.0 (0.0010)          & 399.6  \\
ar9152                    & 837479               & \textbf{837479 (0.0000)}  & \textbf{837479.0 (0.0000)}  & 1285.9 &  & 838752 (0.1520)          & 838752.0 (0.1520)         & 4130.0 &  & 837641 (0.0193)          & 837641.0 (0.0193)          & 3366.7 \\
dga9698                   & 27724                & \textbf{27724 (0.0000)}   & \textbf{27724.0 (0.0000)}   & 844.2  &  & 27735 (0.0397)           & 27735.0 (0.0397)          & 1143.0 &  & \textbf{27724 (0.0000)}  & 27724.5 (0.0018)           & 850.2  \\
ja9847                    & 491924               & \textbf{491924 (0.0000)}  & \textbf{491925.4 (0.0003)}  & 2242.3 &  & 492905 (0.1994)          & 492905.0 (0.1994)         & 4363.2 &  & 492248 (0.0659)          & 492248.0 (0.0659)          & 4461.7 \\
gr9882                    & 300899               & \textbf{300899 (0.0000)}  & \textbf{300900.8 (0.0006)}  & 1362.7 &  & 301094 (0.0648)          & 301095.5 (0.0653)         & 2049.5 &  & 300904 (0.0017)          & 300904.0 (0.0017)          & 2778.0 \\
kz9976                    & 1061881              & \textbf{1061881 (0.0000)} & \textbf{1061881.5 (0.0000)} & 1958.3 &  & 1063701 (0.1714)         & 1063726.0 (0.1737)        & 2243.6 &  & 1061962 (0.0076)         & 1061962.0 (0.0076)         & 2612.1 \\ \hline
Average                   & -                    & \textbf{0.0000}          & \textbf{0.0009}            & 304.4  &  & 0.1344                  & 0.1438                   & 692.5  &  & 0.0861                  & 0.0908                    & 558.0
 \\ \bottomrule
\end{tabular}}
\end{table*}

\subsection{Baseline Algorithms}
For baseline algorithms in the comparison, we choose two state-of-the-art heuristic algorithms, EAX-GA and LKH, as well as one of the state-of-the-art (deep) learning based algorithms, NeuroLKH~\cite{Xin2021}.

For EAX-GA, we generate two baseline algorithms, one with the default parameters (i.e., $N_{pop}=300$, $N_{ch}=30$), so-called EAX-300, the other with a larger population size (i.e., $N_{pop}=400$, $N_{ch}=30$), so-called EAX-400. Note that the termination condition of RHGA, EAX-300, and EAX-400 are the same (see Section \ref{sec_EAX}). We compare RHGA with EAX-400 since their run times for each tested instance are close and a little longer than the run time of EAX-300. Specifically, the average run time for the 138 tested instances of RHGA/EAX-400 is about 37.41\%/38.44\% longer than that of EAX-300. In order to compare the results of EAX-GA and RHGA within the same computation time, it is reasonable to increase the population size of EAX-GA, rather than the cut-off time, since the individuals in the population can hardly be improved after the population converges. 

For the LKH algorithm, we use its newest version\footnote{http://akira.ruc.dk/\%7Ekeld/research/LKH/} as the baseline algorithm. LKH terminates when the number of iterations reaches $n$ (the default termination condition in LKH) or the calculation time exceeds the cut-off time. The cut-off time for LKH is set to be $n/2$ seconds for the instances with less than 70,000 cities, and $n$ seconds for the two super large instances \textit{ch71009} and \textit{pla85900}.

For the NeuroLKH algorithm, we do the comparison with NeuroLKH\_R, which is trained on instances with uniformly distributed nodes, and NeuroLKH\_M, which is trained on a mixture of instances with uniformly distributed nodes, clustered nodes, half uniform and half clustered nodes. The resources required by NeuroLKH are numerous for large scale instances. The performance of NeuroLKH for large instances is also limited due to the small scale of the supervised training instances. Therefore, we only compare RHGA with NeuroLKH on instances with the number of cities ranging from 1,000 to 10,000. Note that in~\cite{Xin2021}, they only reported results on instances with less than 6,000 cities. NeuroLKH terminates when the number of iterations reaches $n$ or the calculation time reaches $n/2$ seconds.

The results of the baseline algorithms 
are all obtained by running their source codes. All the algorithms will terminate their current run when they obtain the known optimum.


\subsection{Comparing RHGA with NeuroLKH}
We first compare RHGA with NeuroLKH\_R and NeuroLKH\_M, in solving all the instances with the number of cities ranging from 1,000 to 10,000 and two-dimensional Euclidean distance (EUC\_2D) metric (NeuroLKH only supports the EUC\_2D metric), a total of 92. We extract the instances that RHGA, NeuroLKH\_R, and NeuroLKH\_M can always yield the optimal solution in each of the 10 runs. The results of the remaining 77 instances are shown in Table \ref{table_NeuroLKH}. We compare the best and average solutions in 10 runs obtained by the algorithms. Column \textit{BKS} indicates the best-known solution of the corresponding instance, and \textit{Time} is the average calculation time (in seconds) of the algorithms. The values in the brackets beside the results equal to the gap of the results to the best-known solutions multiplied by 100. We also provide the average gap of the best and average solutions to the best-known solutions.

From the results in Table \ref{table_NeuroLKH}, we can observe that:

(1) RHGA significantly outperforms NeuroLKH\_R and NeuroLKH\_M. RHGA can yield all the best-known solutions in 10 runs. The best solutions of RHGA are better than those of NeuroLKH\_R (NeuroLKH\_M) in 42 (29) instances. The average solutions of RHGA are better than those of NeuroLKH\_R (NeuroLKH\_M) on 72 (60) instances. The average gaps of the best solutions and average solutions of RHGA are much smaller than those of NeuroLKH\_R and NeuroLKH\_M. The average calculation time of RHGA is also much smaller than that of NeuroLKH\_R and NeuroLKH\_M.

(2) The performance of NeuroLKH\_M is better than that of NeuroLKH\_R, indicating that the performance of NeuroLKH relies on the structure of the training instances. Generating reasonable training instances that help NeuroLKH work well on instances with diverse structures is challenging. Moreover, both NeuroLKH\_R and NeuroLKH\_M are not good at solving large instances, indicating that the bottlenecks in large scale instances still limit algorithms based on deep neural networks.

\begin{sidewaystable*}[thp]
\centering
\caption{Comparison of RHGA and the baseline algorithms, EAX-300, EAX-400, and LKH, on 60 \textit{easy} instances. Best results appear in bold.}
\label{table_Easy}
\scalebox{0.6}{\begin{tabular}{lrrrrrrrrrrrrrrrr} \toprule
\multirow{2}{*}{Instance} & \multirow{2}{*}{BKS} & \multicolumn{3}{c}{RHGA}                                &  & \multicolumn{3}{c}{EAX-300}                             &  & \multicolumn{3}{c}{EAX-400}                             &  & \multicolumn{3}{c}{LKH}                                 \\ \cline{3-5} \cline{7-9} \cline{11-13} \cline{15-17} 
                          &                      & Best (gap\%)               & Average (gap\%)     & Time   &  & Best (gap\%)               & Average (gap\%)     & Time   &  & Best (gap\%)               & Average (gap\%)     & Time   &  & Best (gap\%)               & Average (gap\%)     & Time   \\ \hline
dsj1000                   & 18660188             & \textbf{18660188 (0.0000)} & 18660188.0 (0.0000) & 35.7   &  & \textbf{18660188 (0.0000)} & 18660188.0 (0.0000) & 20.3   &  & \textbf{18660188 (0.0000)} & 18660188.0 (0.0000) & 46.2   &  & \textbf{18660188 (0.0000)} & 18660188.0 (0.0000) & 70.6   \\
pr1002                    & 259045               & \textbf{259045 (0.0000)}   & 259045.0 (0.0000)   & 27.3   &  & \textbf{259045 (0.0000)}   & 259045.0 (0.0000)   & 26.9   &  & \textbf{259045 (0.0000)}   & 259045.0 (0.0000)   & 41.8   &  & \textbf{259045 (0.0000)}   & 259045.6 (0.0002)   & 6.1    \\
u1060                     & 224094               & \textbf{224094 (0.0000)}   & 224094.0 (0.0000)   & 28.4   &  & \textbf{224094 (0.0000)}   & 224094.0 (0.0000)   & 28.8   &  & \textbf{224094 (0.0000)}   & 224094.0 (0.0000)   & 29.3   &  & \textbf{224094 (0.0000)}   & 224107.5 (0.0060)   & 168.9  \\
xit1083                   & 3558                 & \textbf{3558 (0.0000)}     & 3558.0 (0.0000)     & 12.1   &  & \textbf{3558 (0.0000)}     & 3558.0 (0.0000)     & 18.8   &  & \textbf{3558 (0.0000)}     & 3558.0 (0.0000)     & 16.0     &  & \textbf{3558 (0.0000)}     & 3558.0 (0.0000)     & 1.1    \\
vm1084                    & 239297               & \textbf{239297 (0.0000)}   & 239297.0 (0.0000)   & 19.1   &  & \textbf{239297 (0.0000)}   & 239297.0 (0.0000)   & 18.3   &  & \textbf{239297 (0.0000)}   & 239297.0 (0.0000)   & 17.9   &  & \textbf{239297 (0.0000)}   & 239372.6 (0.0316)   & 39.2   \\
pcb1173                   & 56892                & \textbf{56892 (0.0000)}    & 56892.0 (0.0000)    & 21.8   &  & \textbf{56892 (0.0000)}    & 56892.0 (0.0000)    & 36.2   &  & \textbf{56892 (0.0000)}    & 56892.0 (0.0000)    & 38.3   &  & \textbf{56892 (0.0000)}    & 56895.0 (0.0053)    & 7.4    \\
d1291                     & 50801                & \textbf{50801 (0.0000)}    & 50801.0 (0.0000)    & 10.6   &  & \textbf{50801 (0.0000)}    & 50801.0 (0.0000)    & 17.5   &  & \textbf{50801 (0.0000)}    & 50801.0 (0.0000)    & 16.9   &  & \textbf{50801 (0.0000)}    & 50801.0 (0.0000)    & 12.0     \\
rl1304                    & 252948               & \textbf{252948 (0.0000)}   & 252948.0 (0.0000)   & 9.3    &  & \textbf{252948 (0.0000)}   & 252948.0 (0.0000)   & 17.9   &  & \textbf{252948 (0.0000)}   & 252948.0 (0.0000)   & 23.7   &  & \textbf{252948 (0.0000)}   & 253156.4 (0.0824)   & 17.6   \\
rl1323                    & 270199               & \textbf{270199 (0.0000)}   & 270199.0 (0.0000)   & 12.7   &  & \textbf{270199 (0.0000)}   & 270199.0 (0.0000)   & 22.1   &  & \textbf{270199 (0.0000)}   & 270199.0 (0.0000)   & 29.5   &  & \textbf{270199 (0.0000)}   & 270219.6 (0.0076)   & 18.6   \\
dka1376                   & 4666                 & \textbf{4666 (0.0000)}     & 4666.0 (0.0000)     & 17.4   &  & \textbf{4666 (0.0000)}     & 4666.0 (0.0000)     & 27.8   &  & \textbf{4666 (0.0000)}     & 4666.0 (0.0000)     & 37.3   &  & \textbf{4666 (0.0000)}     & 4666.0 (0.0000)     & 1.6    \\
nrw1379                   & 56638                & \textbf{56638 (0.0000)}    & 56638.0 (0.0000)    & 59.9   &  & \textbf{56638 (0.0000)}    & 56638.0 (0.0000)    & 52.1   &  & \textbf{56638 (0.0000)}    & 56638.0 (0.0000)    & 94.3   &  & \textbf{56638 (0.0000)}    & 56640.0 (0.0035)    & 14.9   \\
dca1389                   & 5085                 & \textbf{5085 (0.0000)}     & 5085.0 (0.0000)     & 31.6   &  & \textbf{5085 (0.0000)}     & 5085.0 (0.0000)     & 20.6   &  & \textbf{5085 (0.0000)}     & 5085.0 (0.0000)     & 45.4   &  & \textbf{5085 (0.0000)}     & 5086.4 (0.0275)     & 135.4  \\
fl1400                    & 20127                & \textbf{20127 (0.0000)}    & 20127.0 (0.0000)    & 74.2   &  & \textbf{20127 (0.0000)}    & 20127.0 (0.0000)    & 28.6   &  & \textbf{20127 (0.0000)}    & 20127.0 (0.0000)    & 35.9   &  & 20164 (0.1838)             & 20167.4 (0.2007)    & 703.5  \\
dja1436                   & 5257                 & \textbf{5257 (0.0000)}     & 5257.0 (0.0000)     & 25.2   &  & \textbf{5257 (0.0000)}     & 5257.0 (0.0000)     & 34.1   &  & \textbf{5257 (0.0000)}     & 5257.0 (0.0000)     & 30.7   &  & \textbf{5257 (0.0000)}     & 5257.6 (0.0114)     & 55.7   \\
icw1483                   & 4416                 & \textbf{4416 (0.0000)}     & 4416.0 (0.0000)     & 28.5   &  & \textbf{4416 (0.0000)}     & 4416.0 (0.0000)     & 33.2   &  & \textbf{4416 (0.0000)}     & 4416.0 (0.0000)     & 28.3   &  & \textbf{4416 (0.0000)}     & 4416.0 (0.0000)     & 19.2   \\
fra1488                   & 4264                 & \textbf{4264 (0.0000)}     & 4264.0 (0.0000)     & 18.1   &  & \textbf{4264 (0.0000)}     & 4264.0 (0.0000)     & 36.7   &  & \textbf{4264 (0.0000)}     & 4264.0 (0.0000)     & 46.3   &  & \textbf{4264 (0.0000)}     & 4264.0 (0.0000)     & 1.2    \\
rbv1583                   & 5387                 & \textbf{5387 (0.0000)}     & 5387.0 (0.0000)     & 41.8   &  & \textbf{5387 (0.0000)}     & 5387.0 (0.0000)     & 41.8   &  & \textbf{5387 (0.0000)}     & 5387.0 (0.0000)     & 53.8   &  & \textbf{5387 (0.0000)}     & 5387.1 (0.0019)     & 23.9   \\
rby1599                   & 5533                 & \textbf{5533 (0.0000)}     & 5533.0 (0.0000)     & 37.8   &  & \textbf{5533 (0.0000)}     & 5533.0 (0.0000)     & 33.8   &  & \textbf{5533 (0.0000)}     & 5533.0 (0.0000)     & 62.5   &  & \textbf{5533 (0.0000)}     & 5534.5 (0.0271)     & 75.5   \\
rw1621                    & 26051                & \textbf{26051 (0.0000)}    & 26051.0 (0.0000)    & 47.8   &  & \textbf{26051 (0.0000)}    & 26051.0 (0.0000)    & 35.7   &  & \textbf{26051 (0.0000)}    & 26051.0 (0.0000)    & 46.6   &  & 26053 (0.0077)             & 26076.3 (0.0971)    & 805.4  \\
d1655                     & 62128                & \textbf{62128 (0.0000)}    & 62128.0 (0.0000)    & 47.3   &  & \textbf{62128 (0.0000)}    & 62128.0 (0.0000)    & 29.7   &  & \textbf{62128 (0.0000)}    & 62128.0 (0.0000)    & 57.7   &  & \textbf{62128 (0.0000)}    & 62128.0 (0.0000)    & 8.3    \\
vm1748                    & 336556               & \textbf{336556 (0.0000)}   & 336556.0 (0.0000)   & 53.3   &  & \textbf{336556 (0.0000)}   & 336556.0 (0.0000)   & 43.8   &  & \textbf{336556 (0.0000)}   & 336556.0 (0.0000)   & 65.4   &  & \textbf{336556 (0.0000)}   & 336557.3 (0.0004)   & 20.8   \\
djc1785                   & 6115                 & \textbf{6115 (0.0000)}     & 6115.0 (0.0000)     & 61.0     &  & \textbf{6115 (0.0000)}     & 6115.0 (0.0000)     & 44.4   &  & \textbf{6115 (0.0000)}     & 6115.0 (0.0000)     & 78.9   &  & \textbf{6115 (0.0000)}     & 6115.5 (0.0082)     & 104.1  \\
rl1889                    & 316536               & \textbf{316536 (0.0000)}   & 316536.0 (0.0000)   & 32.7   &  & \textbf{316536 (0.0000)}   & 316536.0 (0.0000)   & 48.1   &  & \textbf{316536 (0.0000)}   & 316536.0 (0.0000)   & 61.6   &  & 316549 (0.0041)            & 316549.8 (0.0044)   & 137.8  \\
dkd1973                   & 6421                 & \textbf{6421 (0.0000)}     & 6421.0 (0.0000)     & 46.2   &  & \textbf{6421 (0.0000)}     & 6421.0 (0.0000)     & 41.8   &  & \textbf{6421 (0.0000)}     & 6421.0 (0.0000)     & 87.3   &  & \textbf{6421 (0.0000)}     & 6421.0 (0.0000)     & 8.0      \\
mu1979                    & 86891                & \textbf{86891 (0.0000)}    & 86891.0 (0.0000)    & 160.4  &  & \textbf{86891 (0.0000)}    & 86891.0 (0.0000)    & 63.4   &  & \textbf{86891 (0.0000)}    & 86891.0 (0.0000)    & 67.6   &  & \textbf{86891 (0.0000)}    & 86892.8 (0.0021)    & 168.0    \\
dcb2086                   & 6600                 & \textbf{6600 (0.0000)}     & 6600.0 (0.0000)     & 64.7   &  & \textbf{6600 (0.0000)}     & 6600.0 (0.0000)     & 69.5   &  & \textbf{6600 (0.0000)}     & 6600.0 (0.0000)     & 80.5   &  & \textbf{6600 (0.0000)}     & 6600.0 (0.0000)     & 36.8   \\
d2103                     & 80450                & \textbf{80450 (0.0000)}    & 80450.0 (0.0000)    & 48.4   &  & \textbf{80450 (0.0000)}    & 80450.0 (0.0000)    & 34.8   &  & \textbf{80450 (0.0000)}    & 80450.0 (0.0000)    & 51.1   &  & 80454 (0.0050)             & 80462.0 (0.0149)    & 164.6  \\
bva2144                   & 6304                 & \textbf{6304 (0.0000)}     & 6304.0 (0.0000)     & 61.0     &  & \textbf{6304 (0.0000)}     & 6304.0 (0.0000)     & 71.3   &  & \textbf{6304 (0.0000)}     & 6304.0 (0.0000)     & 87.4   &  & \textbf{6304 (0.0000)}     & 6304.0 (0.0000)     & 3.9    \\
u2152                     & 64253                & \textbf{64253 (0.0000)}    & 64253.0 (0.0000)    & 61.4   &  & \textbf{64253 (0.0000)}    & 64253.0 (0.0000)    & 62.0     &  & \textbf{64253 (0.0000)}    & 64253.0 (0.0000)    & 81.2   &  & \textbf{64253 (0.0000)}    & 64287.7 (0.0540)    & 135.8  \\
xqc2175                   & 6830                 & \textbf{6830 (0.0000)}     & 6830.0 (0.0000)     & 76.6   &  & \textbf{6830 (0.0000)}     & 6830.0 (0.0000)     & 75.2   &  & \textbf{6830 (0.0000)}     & 6830.0 (0.0000)     & 91.0     &  & \textbf{6830 (0.0000)}     & 6830.5 (0.0073)     & 106.8  \\
ley2323                   & 8352                 & \textbf{8352 (0.0000)}     & 8352.0 (0.0000)     & 49.2   &  & \textbf{8352 (0.0000)}     & 8352.0 (0.0000)     & 52.8   &  & \textbf{8352 (0.0000)}     & 8352.0 (0.0000)     & 45.1   &  & \textbf{8352 (0.0000)}     & 8353.5 (0.0180)     & 86.8   \\
dea2382                   & 8017                 & \textbf{8017 (0.0000)}     & 8017.0 (0.0000)     & 71.6   &  & \textbf{8017 (0.0000)}     & 8017.0 (0.0000)     & 78.4   &  & \textbf{8017 (0.0000)}     & 8017.0 (0.0000)     & 62.2   &  & \textbf{8017 (0.0000)}     & 8017.3 (0.0037)     & 130.0    \\
pr2392                    & 378032               & \textbf{378032 (0.0000)}   & 378032.0 (0.0000)   & 39.1   &  & \textbf{378032 (0.0000)}   & 378032.0 (0.0000)   & 82.7   &  & \textbf{378032 (0.0000)}   & 378032.0 (0.0000)   & 69.3   &  & \textbf{378032 (0.0000)}   & 378032.0 (0.0000)   & 0.6    \\
rbw2481                   & 7724                 & \textbf{7724 (0.0000)}     & 7724.0 (0.0000)     & 59.3   &  & \textbf{7724 (0.0000)}     & 7724.0 (0.0000)     & 67.3   &  & \textbf{7724 (0.0000)}     & 7724.1 (0.0013)     & 99.3   &  & \textbf{7724 (0.0000)}     & 7724.0 (0.0000)     & 3.9    \\
pds2566                   & 7643                 & \textbf{7643 (0.0000)}     & 7643.0 (0.0000)     & 102.6  &  & \textbf{7643 (0.0000)}     & 7643.0 (0.0000)     & 89.4   &  & \textbf{7643 (0.0000)}     & 7643.0 (0.0000)     & 86.6   &  & \textbf{7643 (0.0000)}     & 7643.0 (0.0000)     & 68.3   \\
mlt2597                   & 8071                 & \textbf{8071 (0.0000)}     & 8071.0 (0.0000)     & 47.6   &  & \textbf{8071 (0.0000)}     & 8071.0 (0.0000)     & 91.0     &  & \textbf{8071 (0.0000)}     & 8071.0 (0.0000)     & 117.3  &  & \textbf{8071 (0.0000)}     & 8071.0 (0.0000)     & 16.6   \\
irw2802                   & 8423                 & \textbf{8423 (0.0000)}     & 8423.0 (0.0000)     & 86.1   &  & \textbf{8423 (0.0000)}     & 8423.0 (0.0000)     & 97.9   &  & \textbf{8423 (0.0000)}     & 8423.0 (0.0000)     & 127.7  &  & \textbf{8423 (0.0000)}     & 8424.2 (0.0142)     & 184.9  \\
lsm2854                   & 8014                 & \textbf{8014 (0.0000)}     & 8014.0 (0.0000)     & 103.2  &  & \textbf{8014 (0.0000)}     & 8014.0 (0.0000)     & 104.6  &  & \textbf{8014 (0.0000)}     & 8014.0 (0.0000)     & 137.5  &  & \textbf{8014 (0.0000)}     & 8014.0 (0.0000)     & 103.5  \\
xva2993                   & 8492                 & \textbf{8492 (0.0000)}     & 8492.0 (0.0000)     & 128.1  &  & \textbf{8492 (0.0000)}     & 8492.0 (0.0000)     & 114.5  &  & \textbf{8492 (0.0000)}     & 8492.0 (0.0000)     & 154.9  &  & \textbf{8492 (0.0000)}     & 8493.3 (0.0153)     & 316.0    \\
pcb3038                   & 137694               & \textbf{137694 (0.0000)}   & 137694.0 (0.0000)   & 151.1  &  & \textbf{137694 (0.0000)}   & 137694.0 (0.0000)   & 162.1  &  & \textbf{137694 (0.0000)}   & 137694.0 (0.0000)   & 172.2  &  & \textbf{137694 (0.0000)}   & 137701.2 (0.0052)   & 86.5   \\
dke3097                   & 10539                & \textbf{10539 (0.0000)}    & 10539.0 (0.0000)    & 127.4  &  & \textbf{10539 (0.0000)}    & 10539.0 (0.0000)    & 91.5   &  & \textbf{10539 (0.0000)}    & 10539.0 (0.0000)    & 107.2  &  & \textbf{10539 (0.0000)}    & 10539.1 (0.0009)    & 154.5  \\
lsn3119                   & 9114                 & \textbf{9114 (0.0000)}     & 9114.0 (0.0000)     & 120.3  &  & \textbf{9114 (0.0000)}     & 9114.0 (0.0000)     & 132.5  &  & \textbf{9114 (0.0000)}     & 9114.0 (0.0000)     & 156.9  &  & \textbf{9114 (0.0000)}     & 9114.4 (0.0044)     & 112.9  \\
lta3140                   & 9517                 & \textbf{9517 (0.0000)}     & 9517.0 (0.0000)     & 134.5  &  & \textbf{9517 (0.0000)}     & 9517.0 (0.0000)     & 136.6  &  & \textbf{9517 (0.0000)}     & 9517.1 (0.0011)     & 167.8  &  & \textbf{9517 (0.0000)}     & 9517.7 (0.0074)     & 183.5  \\
dhb3386                   & 11137                & \textbf{11137 (0.0000)}    & 11137.0 (0.0000)    & 127.5  &  & \textbf{11137 (0.0000)}    & 11137.0 (0.0000)    & 154.8  &  & \textbf{11137 (0.0000)}    & 11137.0 (0.0000)    & 183.3  &  & \textbf{11137 (0.0000)}    & 11137.0 (0.0000)    & 101.1  \\
fjr3672                   & 9601                 & \textbf{9601 (0.0000)}     & 9601.0 (0.0000)     & 158.2  &  & \textbf{9601 (0.0000)}     & 9601.0 (0.0000)     & 157.0    &  & \textbf{9601 (0.0000)}     & 9601.0 (0.0000)     & 147.0    &  & \textbf{9601 (0.0000)}     & 9601.0 (0.0000)     & 107.2  \\
ltb3729                   & 11821                & \textbf{11821 (0.0000)}    & 11821.0 (0.0000)    & 171.6  &  & \textbf{11821 (0.0000)}    & 11821.0 (0.0000)    & 165.4  &  & \textbf{11821 (0.0000)}    & 11821.0 (0.0000)    & 237.5  &  & \textbf{11821 (0.0000)}    & 11822.2 (0.0102)    & 716.8  \\
xua3937                   & 11239                & \textbf{11239 (0.0000)}    & 11239.0 (0.0000)    & 134.8  &  & \textbf{11239 (0.0000)}    & 11239.0 (0.0000)    & 147.5  &  & \textbf{11239 (0.0000)}    & 11239.0 (0.0000)    & 197.1  &  & \textbf{11239 (0.0000)}    & 11240.4 (0.0125)    & 639.4  \\
bgb4355                   & 12723                & \textbf{12723 (0.0000)}    & 12723.0 (0.0000)    & 238.1  &  & \textbf{12723 (0.0000)}    & 12723.0 (0.0000)    & 176.9  &  & \textbf{12723 (0.0000)}    & 12723.0 (0.0000)    & 224.7  &  & \textbf{12723 (0.0000)}    & 12728.0 (0.0393)    & 582.5  \\
bgd4396                   & 13009                & \textbf{13009 (0.0000)}    & 13009.0 (0.0000)    & 205.0    &  & \textbf{13009 (0.0000)}    & 13009.0 (0.0000)    & 145.3  &  & \textbf{13009 (0.0000)}    & 13009.1 (0.0008)    & 216.1  &  & \textbf{13009 (0.0000)}    & 13010.3 (0.0100)    & 388.3  \\
frv4410                   & 10711                & \textbf{10711 (0.0000)}    & 10711.0 (0.0000)    & 224.7  &  & \textbf{10711 (0.0000)}    & 10711.0 (0.0000)    & 201.4  &  & \textbf{10711 (0.0000)}    & 10711.0 (0.0000)    & 266.7  &  & \textbf{10711 (0.0000)}    & 10712.1 (0.0103)    & 356.0    \\
fnl4461                   & 182566               & \textbf{182566 (0.0000)}   & 182566.0 (0.0000)   & 497.9  &  & \textbf{182566 (0.0000)}   & 182566.0 (0.0000)   & 628.5  &  & \textbf{182566 (0.0000)}   & 182566.4 (0.0002)   & 583.6  &  & \textbf{182566 (0.0000)}   & 182566.5 (0.0003)   & 38.0     \\
bgf4475                   & 13221                & \textbf{13221 (0.0000)}    & 13221.0 (0.0000)    & 222.7  &  & \textbf{13221 (0.0000)}    & 13221.0 (0.0000)    & 226.1  &  & \textbf{13221 (0.0000)}    & 13221.1 (0.0008)    & 204.7  &  & \textbf{13221 (0.0000)}    & 13224.4 (0.0257)    & 763.3  \\
ca4663                    & 1290319              & \textbf{1290319 (0.0000)}  & 1290319.0 (0.0000)  & 395.1  &  & \textbf{1290319 (0.0000)}  & 1290319.0 (0.0000)  & 372.2  &  & \textbf{1290319 (0.0000)}  & 1290319.0 (0.0000)  & 529.3  &  & \textbf{1290319 (0.0000)}  & 1290338.7 (0.0015)  & 508.0    \\
xqd4966                   & 15316                & \textbf{15316 (0.0000)}    & 15316.0 (0.0000)    & 325.7  &  & \textbf{15316 (0.0000)}    & 15316.0 (0.0000)    & 306.3  &  & \textbf{15316 (0.0000)}    & 15316.0 (0.0000)    & 287.6  &  & \textbf{15316 (0.0000)}    & 15316.2 (0.0013)    & 724.0    \\
fea5557                   & 15445                & \textbf{15445 (0.0000)}    & 15445.0 (0.0000)    & 293.8  &  & \textbf{15445 (0.0000)}    & 15445.0 (0.0000)    & 231.1  &  & \textbf{15445 (0.0000)}    & 15445.0 (0.0000)    & 265.3  &  & \textbf{15445 (0.0000)}    & 15445.8 (0.0052)    & 521.2  \\
rl5915                    & 565530               & \textbf{565530 (0.0000)}   & 565530.0 (0.0000)   & 289.7  &  & \textbf{565530 (0.0000)}   & 565530.0 (0.0000)   & 330.7  &  & \textbf{565530 (0.0000)}   & 565530.0 (0.0000)   & 468.7  &  & 565544 (0.0025)            & 565581.2 (0.0091)   & 412.9  \\
bnd7168                   & 21834                & \textbf{21834 (0.0000)}    & 21834.0 (0.0000)    & 396.6  &  & \textbf{21834 (0.0000)}    & 21834.0 (0.0000)    & 406.3  &  & \textbf{21834 (0.0000)}    & 21834.0 (0.0000)    & 494.4  &  & \textbf{21834 (0.0000)}    & 21834.5 (0.0023)    & 262.3  \\
ym7663                    & 238314               & \textbf{238314 (0.0000)}   & 238314.0 (0.0000)   & 1071.9 &  & \textbf{238314 (0.0000)}   & 238314.0 (0.0000)   & 833.7  &  & \textbf{238314 (0.0000)}   & 238314.1 (0.0000)   & 1053.5 &  & \textbf{238314 (0.0000)}   & 238318.4 (0.0018)   & 975.1  \\
dga9698                   & 27724                & \textbf{27724 (0.0000)}    & 27724.0 (0.0000)    & 844.2  &  & \textbf{27724 (0.0000)}    & 27724.0 (0.0000)    & 531.0    &  & \textbf{27724 (0.0000)}    & 27724.1 (0.0004)    & 808.1  &  & \textbf{27724 (0.0000)}    & 27726.7 (0.0097)    & 3073.4 \\
brd14051                  & 469385               & \textbf{469385 (0.0000)}   & 469385.0 (0.0000)   & 5510.8 &  & \textbf{469385 (0.0000)}   & 469385.0 (0.0000)   & 4499.1 &  & \textbf{469385 (0.0000)}   & 469385.0 (0.0000)   & 5925.2 &  & 469389 (0.0009)            & 469393.4 (0.0018)   & 5997.3 \\ \hline
Average                   &  --                    & \textbf{0.0000}           & \textbf{0.0000}    & 226.7  &  & \textbf{0.0000}           & \textbf{0.0000}    & 199.2  &  & \textbf{0.0000}           & 0.0001             & 252.5  &  & 0.0034                    & 0.0134             & 344.8 \\ \bottomrule
\end{tabular}}
\end{sidewaystable*}

\begin{sidewaystable*}[thp]
\centering
\caption{Comparison of RHGA and the baseline algorithms, EAX-300, EAX-400, and LKH, on 62 \textit{medium} instances. Best results appear in bold.}
\label{table_Medium}
\scalebox{0.6}{\begin{tabular}{lrrrrrrrrrrrrrrrr} \toprule
\multirow{2}{*}{Instance} & \multirow{2}{*}{BKS} & \multicolumn{3}{c}{RHGA}                                 &  & \multicolumn{3}{c}{EAX-300}                              &  & \multicolumn{3}{c}{EAX-400}                              &  & \multicolumn{3}{c}{LKH}                                  \\ \cline{3-5} \cline{7-9} \cline{11-13} \cline{15-17} 
                          &                      & Best (gap\%)               & Average (gap\%)     & Time    &  & Best (gap\%)               & Average (gap\%)     & Time    &  & Best (gap\%)               & Average (gap\%)     & Time    &  & Best (gap\%)               & Average (gap\%)     & Time    \\ \hline
si1032                    & 92650                & \textbf{92650 (0.0000)}    & 92650.1 (0.0001)    & 8.1     &  & \textbf{92650 (0.0000)}    & 92650.5 (0.0005)    & 1.6     &  & \textbf{92650 (0.0000)}    & 92650.1 (0.0001)    & 3.7     &  & \textbf{92650 (0.0000)}    & 92650.0 (0.0000)    & 34.4    \\
u1432                     & 152970               & \textbf{152970 (0.0000)}   & 152970.0 (0.0000)   & 25.7    &  & \textbf{152970 (0.0000)}   & 152973.6 (0.0024)   & 38.3    &  & \textbf{152970 (0.0000)}   & 152979.0 (0.0059)   & 56.3    &  & \textbf{152970 (0.0000)}   & 152970.0 (0.0000)   & 0.9     \\
fl1577                    & 22249                & \textbf{22249 (0.0000)}    & 22249.0 (0.0000)    & 67.6    &  & \textbf{22249 (0.0000)}    & 22309.5 (0.2719)    & 38.9    &  & \textbf{22249 (0.0000)}    & 22322.9 (0.3321)    & 34.3    &  & 22261 (0.0539)             & 22262.1 (0.0589)    & 765.1   \\
fnb1615                   & 4956                 & \textbf{4956 (0.0000)}     & 4956.1 (0.0020)     & 48.4    &  & \textbf{4956 (0.0000)}     & 4956.6 (0.0121)     & 35.9    &  & \textbf{4956 (0.0000)}     & 4956.7 (0.0141)     & 77.6    &  & \textbf{4956 (0.0000)}     & 4956.0 (0.0000)     & 36.5    \\
u1817                     & 57201                & \textbf{57201 (0.0000)}    & 57209.1 (0.0142)    & 51.0      &  & \textbf{57201 (0.0000)}    & 57216.0 (0.0262)    & 67.5    &  & \textbf{57201 (0.0000)}    & 57216.6 (0.0273)    & 55.5    &  & \textbf{57201 (0.0000)}    & 57251.1 (0.0876)    & 93.7    \\
dcc1911                   & 6396                 & \textbf{6396 (0.0000)}     & 6396.0 (0.0000)     & 53.7    &  & \textbf{6396 (0.0000)}     & 6396.1 (0.0016)     & 65.6    &  & \textbf{6396 (0.0000)}     & 6396.0 (0.0000)     & 67.5    &  & \textbf{6396 (0.0000)}     & 6397.0 (0.0156)     & 173.6   \\
djb2036                   & 6197                 & \textbf{6197 (0.0000)}     & 6197.0 (0.0000)     & 55.5    &  & \textbf{6197 (0.0000)}     & 6197.2 (0.0032)     & 76.6    &  & \textbf{6197 (0.0000)}     & 6197.0 (0.0000)     & 86.0      &  & \textbf{6197 (0.0000)}     & 6197.1 (0.0016)     & 26.0      \\
bck2217                   & 6764                 & \textbf{6764 (0.0000)}     & 6764.3 (0.0044)     & 81.9    &  & \textbf{6764 (0.0000)}     & 6764.7 (0.0103)     & 104.3   &  & \textbf{6764 (0.0000)}     & 6764.2 (0.0030)     & 121.4   &  & 6765 (0.0148)              & 6765.2 (0.0177)     & 221.2   \\
xpr2308                   & 7219                 & \textbf{7219 (0.0000)}     & 7219.1 (0.0014)     & 80.4    &  & \textbf{7219 (0.0000)}     & 7219.2 (0.0028)     & 98.4    &  & \textbf{7219 (0.0000)}     & 7219.0 (0.0000)     & 68.8    &  & \textbf{7219 (0.0000)}     & 7219.2 (0.0028)     & 44.3    \\
bch2762                   & 8234                 & \textbf{8234 (0.0000)}     & 8234.1 (0.0012)     & 131.6   &  & \textbf{8234 (0.0000)}     & 8234.4 (0.0049)     & 129.8   &  & \textbf{8234 (0.0000)}     & 8234.0 (0.0000)     & 132.6   &  & \textbf{8234 (0.0000)}     & 8235.1 (0.0134)     & 249.9   \\
dbj2924                   & 10128                & \textbf{10128 (0.0000)}    & 10128.0 (0.0000)    & 127.1   &  & \textbf{10128 (0.0000)}    & 10128.1 (0.0010)    & 127.9   &  & \textbf{10128 (0.0000)}    & 10128.0 (0.0000)    & 99.1    &  & \textbf{10128 (0.0000)}    & 10128.1 (0.0010)    & 120.0     \\
pia3056                   & 8258                 & \textbf{8258 (0.0000)}     & 8258.5 (0.0061)     & 156.4   &  & \textbf{8258 (0.0000)}     & 8258.8 (0.0097)     & 141.6   &  & \textbf{8258 (0.0000)}     & 8258.6 (0.0073)     & 157.1   &  & \textbf{8258 (0.0000)}     & 8262.2 (0.0509)     & 526.0     \\
fdp3256                   & 10008                & \textbf{10008 (0.0000)}    & 10008.1 (0.0010)    & 127.4   &  & \textbf{10008 (0.0000)}    & 10008.2 (0.0020)    & 157.8   &  & \textbf{10008 (0.0000)}    & 10008.5 (0.0050)    & 206.2   &  & 10009 (0.0100)             & 10009.7 (0.0170)    & 450.9   \\
beg3293                   & 9772                 & \textbf{9772 (0.0000)}     & 9772.2 (0.0020)     & 131.1   &  & \textbf{9772 (0.0000)}     & 9772.0 (0.0000)     & 142.3   &  & \textbf{9772 (0.0000)}     & 9772.1 (0.0010)     & 182.6   &  & \textbf{9772 (0.0000)}     & 9772.2 (0.0020)     & 171.3   \\
nu3496                    & 96132                & \textbf{96132 (0.0000)}    & 96132.1 (0.0001)    & 169.1   &  & \textbf{96132 (0.0000)}    & 96132.0 (0.0000)    & 150.8   &  & \textbf{96132 (0.0000)}    & 96132.0 (0.0000)    & 189.0     &  & 96180 (0.0499)             & 96201.4 (0.0722)    & 1749.1  \\
fjs3649                   & 9272                 & \textbf{9272 (0.0000)}     & 9272.0 (0.0000)     & 176.3   &  & \textbf{9272 (0.0000)}     & 9272.5 (0.0054)     & 167.4   &  & \textbf{9272 (0.0000)}     & 9272.0 (0.0000)     & 189.4   &  & \textbf{9272 (0.0000)}     & 9272.0 (0.0000)     & 88.5    \\
dlb3694                   & 10959                & \textbf{10959 (0.0000)}    & 10959.3 (0.0027)    & 216.7   &  & \textbf{10959 (0.0000)}    & 10959.3 (0.0027)    & 185.4   &  & \textbf{10959 (0.0000)}    & 10959.1 (0.0009)    & 184.0     &  & \textbf{10959 (0.0000)}    & 10959.7 (0.0064)    & 334.4   \\
xqe3891                   & 11995                & \textbf{11995 (0.0000)}    & 11996.1 (0.0092)    & 228.8   &  & \textbf{11995 (0.0000)}    & 11996.0 (0.0083)    & 245.7   &  & \textbf{11995 (0.0000)}    & 11995.9 (0.0075)    & 347.4   &  & \textbf{11995 (0.0000)}    & 11998.2 (0.0267)    & 454.0     \\
dkc3938                   & 12503                & \textbf{12503 (0.0000)}    & 12503.0 (0.0000)    & 187.2   &  & \textbf{12503 (0.0000)}    & 12503.1 (0.0008)    & 140.9   &  & \textbf{12503 (0.0000)}    & 12503.0 (0.0000)    & 236.5   &  & \textbf{12503 (0.0000)}    & 12503.8 (0.0064)    & 697.1   \\
dkf3954                   & 12538                & \textbf{12538 (0.0000)}    & 12538.0 (0.0000)    & 180.3   &  & \textbf{12538 (0.0000)}    & 12538.2 (0.0016)    & 124.4   &  & \textbf{12538 (0.0000)}    & 12538.0 (0.0000)    & 227.3   &  & \textbf{12538 (0.0000)}    & 12538.6 (0.0048)    & 113.9   \\
fqm5087                   & 13029                & \textbf{13029 (0.0000)}    & 13029.0 (0.0000)    & 428.4   &  & \textbf{13029 (0.0000)}    & 13029.1 (0.0008)    & 337.7   &  & \textbf{13029 (0.0000)}    & 13029.1 (0.0008)    & 400.0     &  & \textbf{13029 (0.0000)}    & 13029.5 (0.0038)    & 1790.7  \\
rl5934                    & 556045               & \textbf{556045 (0.0000)}   & 556072.3 (0.0049)   & 476.1   &  & \textbf{556045 (0.0000)}   & 556090.9 (0.0083)   & 311.0     &  & \textbf{556045 (0.0000)}   & 556054.1 (0.0016)   & 474.4   &  & 556136 (0.0164)            & 556309.8 (0.0476)   & 734.2   \\
tz6117                    & 394718               & \textbf{394718 (0.0000)}   & 394721.2 (0.0008)   & 694.0     &  & \textbf{394718 (0.0000)}   & 394721.6 (0.0009)   & 572.0     &  & \textbf{394718 (0.0000)}   & 394721.4 (0.0009)   & 829.8   &  & 394726 (0.0020)            & 394747.6 (0.0075)   & 3068.4  \\
xsc6880                   & 21535                & \textbf{21535 (0.0000)}    & 21535.1 (0.0005)    & 641.7   &  & \textbf{21535 (0.0000)}    & 21535.6 (0.0028)    & 454.3   &  & \textbf{21535 (0.0000)}    & 21535.1 (0.0005)    & 762.6   &  & 21537 (0.0093)             & 21540.6 (0.0260)    & 1743.2  \\
eg7146                    & 172386               & \textbf{172386 (0.0000)}   & 172386.0 (0.0000)   & 1199.0    &  & \textbf{172386 (0.0000)}   & 172386.2 (0.0001)   & 1058.4  &  & \textbf{172386 (0.0000)}   & 172386.0 (0.0000)   & 981.6   &  & 172738 (0.2042)            & 172738.7 (0.2046)   & 2068.6  \\
lap7454                   & 19535                & \textbf{19535 (0.0000)}    & 19535.0 (0.0000)    & 457.2   &  & \textbf{19535 (0.0000)}    & 19535.1 (0.0005)    & 498.9   &  & \textbf{19535 (0.0000)}    & 19535.0 (0.0000)    & 496.2   &  & \textbf{19535 (0.0000)}    & 19537.1 (0.0107)    & 1283.5  \\
pm8079                    & 114855               & \textbf{114855 (0.0000)}   & 114855.3 (0.0003)   & 1301.2  &  & \textbf{114855 (0.0000)}   & 114884.9 (0.0260)   & 1550.1  &  & \textbf{114855 (0.0000)}   & 114884.0 (0.0252)   & 2081.2  &  & 114872 (0.0148)            & 114893.9 (0.0339)   & 4044.8  \\
ida8197                   & 22338                & \textbf{22338 (0.0000)}    & 22338.1 (0.0004)    & 424.8   &  & \textbf{22338 (0.0000)}    & 22338.2 (0.0009)    & 501.7   &  & \textbf{22338 (0.0000)}    & 22338.0 (0.0000)    & 740.8   &  & \textbf{22338 (0.0000)}    & 22339.2 (0.0054)    & 930.7   \\
ei8246                    & 206171               & \textbf{206171 (0.0000)}   & 206171.6 (0.0003)   & 1346.3  &  & \textbf{206171 (0.0000)}   & 206172.7 (0.0008)   & 1295.8  &  & \textbf{206171 (0.0000)}   & 206171.0 (0.0000)   & 1943.4  &  & \textbf{206171 (0.0000)}   & 206175.2 (0.0020)   & 1213.8  \\
ar9152                    & 837479               & \textbf{837479 (0.0000)}   & 837479.0 (0.0000)   & 1285.9  &  & \textbf{837479 (0.0000)}   & 837528.0 (0.0059)   & 1202.9  &  & \textbf{837479 (0.0000)}   & 837554.2 (0.0090)   & 1696.5  &  & 837575 (0.0115)            & 837641.8 (0.0194)   & 4579.3  \\
ja9847                    & 491924               & \textbf{491924 (0.0000)}   & 491925.4 (0.0003)   & 2242.3  &  & \textbf{491924 (0.0000)}   & 491927.4 (0.0007)   & 961.7   &  & \textbf{491924 (0.0000)}   & 491926.6 (0.0005)   & 1430.9  &  & 491947 (0.0047)            & 492073.2 (0.0303)   & 2859.2  \\
gr9882                    & 300899               & \textbf{300899 (0.0000)}   & 300900.8 (0.0006)   & 1362.7  &  & \textbf{300899 (0.0000)}   & 300901.6 (0.0009)   & 1316.7  &  & \textbf{300899 (0.0000)}   & 300900.4 (0.0005)   & 1396.9  &  & \textbf{300899 (0.0000)}   & 300901.0 (0.0007)   & 1457.0    \\
kz9976                    & 1061881              & \textbf{1061881 (0.0000)}  & 1061881.5 (0.0000)  & 1958.3  &  & \textbf{1061881 (0.0000)}  & 1061882.0 (0.0001)  & 1745.1  &  & \textbf{1061881 (0.0000)}  & 1061881.0 (0.0000)  & 1725.6  &  & \textbf{1061881 (0.0000)}  & 1061941.2 (0.0057)  & 2113.4  \\
xmc10150                  & 28387                & \textbf{28387 (0.0000)}    & 28387.1 (0.0004)    & 988.6   &  & \textbf{28387 (0.0000)}    & 28387.3 (0.0011)    & 688.0     &  & \textbf{28387 (0.0000)}    & 28387.3 (0.0011)    & 951.5   &  & \textbf{28387 (0.0000)}    & 28389.3 (0.0081)    & 2025.7  \\
fi10639                   & 520527               & \textbf{520527 (0.0000)}   & 520527.1 (0.0000)   & 2336.6  &  & \textbf{520527 (0.0000)}   & 520527.3 (0.0001)   & 1847.9  &  & \textbf{520527 (0.0000)}   & 520527.1 (0.0000)   & 2340.6  &  & \textbf{520531 (0.0008)}   & 520561.8 (0.0067)   & 3260.7  \\
rl11849                   & 923288               & \textbf{923288 (0.0000)}   & 923288.0 (0.0000)   & 2383.0    &  & \textbf{923288 (0.0000)}   & 923291.6 (0.0004)   & 1729.9  &  & \textbf{923288 (0.0000)}   & 923288.0 (0.0000)   & 2173.4  &  & \textbf{923288 (0.0000)}   & 923362.7 (0.0081)   & 3684.3  \\
usa13509                  & 19982859             & \textbf{19982859 (0.0000)} & 19982881.7 (0.0001) & 5056.8  &  & \textbf{19982859 (0.0000)} & 19982894.6 (0.0002) & 3357.8  &  & \textbf{19982859 (0.0000)} & 19982859.0 (0.0000) & 4688.6  &  & \textbf{19982859 (0.0000)} & 19983103.4 (0.0012) & 5146.7  \\
xvb13584                  & 37083                & \textbf{37083 (0.0000)}    & 37083.2 (0.0005)    & 1821.8  &  & \textbf{37083 (0.0000)}    & 37083.2 (0.0005)    & 1691.1  &  & \textbf{37083 (0.0000)}    & 37083.0 (0.0000)    & 2179.5  &  & \textbf{37083 (0.0000)}    & 37088.7 (0.0154)    & 4700.7  \\
mo14185                   & 427377               & \textbf{427377 (0.0000)}   & 427377.6 (0.0001)   & 4215.7  &  & \textbf{427377 (0.0000)}   & 427378.2 (0.0003)   & 2643.3  &  & \textbf{427377 (0.0000)}   & 427377.2 (0.0000)   & 4121.8  &  & 427382 (0.0012)            & 427399.5 (0.0053)   & 6264.8  \\
xrb14233                  & 45462                & \textbf{45462 (0.0000)}    & 45463.3 (0.0029)    & 2122.4  &  & \textbf{45462 (0.0000)}    & 45463.6 (0.0035)    & 1518.4  &  & \textbf{45462 (0.0000)}    & 45463.6 (0.0035)    & 1988.5  &  & 45464 (0.0044)             & 45468.5 (0.0143)    & 5723.8  \\
ho14473                   & 177092               & \textbf{177092 (0.0000)}   & 177092.2 (0.0001)   & 6841.0    &  & \textbf{177092 (0.0000)}   & 177092.4 (0.0002)   & 4906.4  &  & \textbf{177092 (0.0000)}   & 177092.3 (0.0002)   & 5930.9  &  & 177235 (0.0807)            & 177303.2 (0.1193)   & 7339.7  \\
d15112                    & 1573084              & \textbf{1573084 (0.0000)}  & 1573084.2 (0.0000)  & 6941.3  &  & \textbf{1573084 (0.0000)}  & 1573084.8 (0.0001)  & 5947.6  &  & \textbf{1573084 (0.0000)}  & 1573084.6 (0.0000)  & 8157.7  &  & 1573085 (0.0001)           & 1573146.9 (0.0040)  & 7257.4  \\
it16862                   & 557315               & \textbf{557315 (0.0000)}   & 557316.9 (0.0003)   & 7958.8  &  & \textbf{557315 (0.0000)}   & 557319.3 (0.0008)   & 3518.9  &  & \textbf{557315 (0.0000)}   & 557317.6 (0.0005)   & 5965.3  &  & 557321 (0.0011)            & 557343.0 (0.0050)   & 8436.6  \\
xia16928                  & 52850                & \textbf{52850 (0.0000)}    & 52850.2 (0.0004)    & 3124.1  &  & \textbf{52850 (0.0000)}    & 52850.7 (0.0013)    & 1897.0    &  & \textbf{52850 (0.0000)}    & 52850.7 (0.0013)    & 3283.7  &  & \textbf{52850 (0.0000)}    & 52856.3 (0.0119)    & 8175.8  \\
pjh17845                  & 48092                & \textbf{48092 (0.0000)}    & 48092.5 (0.0010)    & 3292.7  &  & \textbf{48092 (0.0000)}    & 48092.2 (0.0004)    & 2449.1  &  & \textbf{48092 (0.0000)}    & 48092.6 (0.0012)    & 3608.3  &  & 48094 (0.0042)             & 48100.6 (0.0179)    & 8769.8  \\
d18512                    & 645238               & \textbf{645238 (0.0000)}   & 645238.6 (0.0001)   & 9809.9  &  & \textbf{645238 (0.0000)}   & 645238.7 (0.0001)   & 10085.7 &  & \textbf{645238 (0.0000)}   & 645238.2 (0.0000)   & 9624.3  &  & 645241 (0.0005)            & 645253.4 (0.0024)   & 9256.5  \\
frh19289                  & 55798                & \textbf{55798 (0.0000)}    & 55798.0 (0.0000)    & 3064.7  &  & \textbf{55798 (0.0000)}    & 55798.2 (0.0004)    & 3174.6  &  & \textbf{55798 (0.0000)}    & 55798.0 (0.0000)    & 3562.2  &  & 55800 (0.0036)             & 55805.9 (0.0142)    & 8568.0    \\
fnc19402                  & 59287                & \textbf{59287 (0.0000)}    & 59287.6 (0.0010)    & 2729.1  &  & \textbf{59287 (0.0000)}    & 59287.6 (0.0010)    & 2296.3  &  & \textbf{59287 (0.0000)}    & 59287.1 (0.0002)    & 2698.1  &  & 59290 (0.0051)             & 59301.1 (0.0238)    & 9228.0    \\
ido21215                  & 63517                & \textbf{63517 (0.0000)}    & 63517.1 (0.0002)    & 4824.3  &  & \textbf{63517 (0.0000)}    & 63517.5 (0.0008)    & 3246.1  &  & \textbf{63517 (0.0000)}    & 63517.1 (0.0002)    & 5377.1  &  & 63521 (0.0063)             & 63529.8 (0.0202)    & 10597.6 \\
fma21553                  & 66527                & \textbf{66527 (0.0000)}    & 66527.6 (0.0009)    & 3785.4  &  & \textbf{66527 (0.0000)}    & 66527.8 (0.0012)    & 2577.5  &  & \textbf{66527 (0.0000)}    & 66527.5 (0.0008)    & 4394.4  &  & 66529 (0.0030)             & 66536.9 (0.0149)    & 10776.3 \\
lsb22777                  & 60977                & \textbf{60977 (0.0000)}    & 60977.0 (0.0000)    & 5064.3  &  & \textbf{60977 (0.0000)}    & 60977.7 (0.0011)    & 3160.8  &  & \textbf{60977 (0.0000)}    & 60977.1 (0.0002)    & 5097.9  &  & 60981 (0.0066)             & 60992.4 (0.0253)    & 11388.5 \\
xrh24104                  & 69294                & \textbf{69294 (0.0000)}    & 69294.0 (0.0000)    & 5711.5  &  & \textbf{69294 (0.0000)}    & 69294.1 (0.0001)    & 3540.6  &  & \textbf{69294 (0.0000)}    & 69294.1 (0.0001)    & 6528.5  &  & 69298 (0.0058)             & 69306.8 (0.0185)    & 12052.6 \\
sw24978                   & 855597               & \textbf{855597 (0.0000)}   & 855599.9 (0.0003)   & 9211.3  &  & \textbf{855597 (0.0000)}   & 855599.6 (0.0003)   & 6806.4  &  & \textbf{855597 (0.0000)}   & 855598.8 (0.0002)   & 10893.9 &  & \textbf{855597 (0.0000)}   & 855636.3 (0.0046)   & 12489.3 \\
bbz25234                  & 69335                & \textbf{69335 (0.0000)}    & 69335.0 (0.0000)    & 5939.0    &  & \textbf{69335 (0.0000)}    & 69335.7 (0.0010)    & 3990.1  &  & \textbf{69335 (0.0000)}    & 69335.3 (0.0004)    & 6340.2  &  & 69341 (0.0087)             & 69350.7 (0.0226)    & 12617.5 \\
irx28268                  & 72607                & \textbf{72607 (0.0000)}    & 72607.0 (0.0000)    & 7975.4  &  & \textbf{72607 (0.0000)}    & 72607.2 (0.0003)    & 4967.4  &  & \textbf{72607 (0.0000)}    & 72607.0 (0.0000)    & 8187.6  &  & 72611 (0.0055)             & 72623.1 (0.0222)    & 14134.8 \\
fyg28534                  & 78562                & \textbf{78562 (0.0000)}    & 78562.3 (0.0004)    & 8673.2  &  & \textbf{78562 (0.0000)}    & 78562.5 (0.0006)    & 6067.6  &  & \textbf{78562 (0.0000)}    & 78562.4 (0.0005)    & 8836.0    &  & 78567 (0.0064)             & 78572.6 (0.0135)    & 14268.1 \\
boa28924                  & 79622                & \textbf{79622 (0.0000)}    & 79623.2 (0.0015)    & 7579.2  &  & \textbf{79622 (0.0000)}    & 79623.2 (0.0015)    & 5828.6  &  & \textbf{79622 (0.0000)}    & 79623.0 (0.0013)    & 7819.9  &  & 79630 (0.0100)             & 79635.0 (0.0163)    & 14462.6 \\
ird29514                  & 80353                & \textbf{80353 (0.0000)}    & 80353.9 (0.0011)    & 7170.3  &  & \textbf{80353 (0.0000)}    & 80354.2 (0.0015)    & 4730.5  &  & \textbf{80353 (0.0000)}    & 80353.7 (0.0009)    & 6238.8  &  & 80361 (0.0100)             & 80369.4 (0.0204)    & 14757.3 \\
pbh30440                  & 88313                & \textbf{88313 (0.0000)}    & 88313.2 (0.0002)    & 7304.5  &  & \textbf{88313 (0.0000)}    & 88313.7 (0.0008)    & 5315.9  &  & \textbf{88313 (0.0000)}    & 88313.2 (0.0002)    & 7600.3  &  & 88314 (0.0011)             & 88324.7 (0.0132)    & 15220.5 \\
fry33203                  & 97240                & \textbf{97240 (0.0000)}    & 97241.3 (0.0013)    & 8381.0    &  & \textbf{97240 (0.0000)}    & 97241.2 (0.0012)    & 6168.5  &  & \textbf{97240 (0.0000)}    & 97240.5 (0.0005)    & 8271.4  &  & 97247 (0.0072)             & 97259.8 (0.0204)    & 16601.7 \\
bby34656                  & 99159                & \textbf{99159 (0.0000)}    & 99159.7 (0.0007)    & 14413.7 &  & \textbf{99159 (0.0000)}    & 99160.5 (0.0015)    & 9448.3  &  & \textbf{99159 (0.0000)}    & 99159.7 (0.0007)    & 12360.9 &  & 99169 (0.0101)             & 99179.3 (0.0205)    & 17328.7 \\
rbz43748                  & 125183               & \textbf{125183 (0.0000)}   & 125183.7 (0.0006)   & 16821.3 &  & \textbf{125183 (0.0000)}   & 125184.2 (0.0010)   & 13560.3 &  & \textbf{125183 (0.0000)}   & 125183.9 (0.0007)   & 18505.8 &  & 125199 (0.0128)            & 125209.6 (0.0212)   & 21874.8 \\ \hline
Average                   &  --                    & \textbf{0.0000}           & \textbf{0.0011}    & 3091.3  &  & \textbf{0.0000}           & 0.0071             & 2277.5  &  & \textbf{0.0000}           & 0.0074             & 3151.8  &  & 0.0095                    & 0.0209             & 5333.0  \\ \bottomrule
\end{tabular}}
\end{sidewaystable*}

\begin{sidewaystable}[thp]
\centering
\caption{Comparison of RHGA and the baseline algorithms, EAX-300, EAX-400, and LKH, on 16 \textit{hard} instances. Best results appear in bold.}
\label{table_Hard}
\scalebox{0.61}{\begin{tabular}{lrrrrrrrrrrrrrrrr} \toprule
\multirow{2}{*}{Instance} & \multirow{2}{*}{BKS} & \multicolumn{3}{c}{RHGA}                                    &  & \multicolumn{3}{c}{EAX-300}                              &  & \multicolumn{3}{c}{EAX-400}                               &  & \multicolumn{3}{c}{LKH}                                   \\ \cline{3-5} \cline{7-9} \cline{11-13} \cline{15-17} 
                          &                      & Best (gap\%)                & Average (gap\%)      & Time     &  & Best (gap\%)             & Average (gap\%)      & Time     &  & Best (gap\%)              & Average (gap\%)      & Time     &  & Best (gap\%)               & Average (gap\%)      & Time    \\ \hline
u2319                     & 234256               & \textbf{234256 (0.0000)}    & 234256.0 (0.0000)    & 99.1     &  & 234273 (0.0073)          & 234330.4 (0.0318)    & 156.5    &  & \textbf{234256 (0.0000)}  & 234318.8 (0.0268)    & 136.9    &  & \textbf{234256 (0.0000)}   & 234256.0 (0.0000)    & 1.1     \\
fl3795                    & 28772                & \textbf{28772 (0.0000)}     & 28777.6 (0.0195)     & 428.3    &  & 28815 (0.1495)           & 28824.0 (0.1807)     & 140.6    &  & 28779 (0.0243)            & 28821.5 (0.1720)     & 202.9    &  & 28819 (0.1634)             & 28906.5 (0.4675)     & 1906.5  \\
pla7397                   & 23260728             & \textbf{23260728 (0.0000)}  & 23260805.4 (0.0003)  & 791.8    &  & 23260814 (0.0004)        & 23261302.6 (0.0025)  & 587.5    &  & 23260814 (0.0004)         & 23261052.6 (0.0014)  & 624.4    &  & \textbf{23260728 (0.0000)} & 23260728.0 (0.0000)  & 330.2   \\
vm22775                   & 569288               & \textbf{569288 (0.0000)}    & 569291.6 (0.0006)    & 9692.5   &  & 569289 (0.0002)          & 569293.6 (0.0010)    & 5414.7   &  & \textbf{569288 (0.0000)}  & 569291.3 (0.0006)    & 9528.0     &  & 569298 (0.0018)            & 569317.7 (0.0052)    & 11389.7 \\
icx28698                  & 78087                & \textbf{78088 (0.0013)}     & 78089.1 (0.0027)     & 8425.1   &  & 78089 (0.0026)           & 78090.1 (0.0040)     & 6000.7   &  & 78089 (0.0026)            & 78089.1 (0.0027)     & 9266.9   &  & 78098 (0.0141)             & 78106.1 (0.0245)     & 14349.4 \\
xib32892                  & 96757                & \textbf{96757 (0.0000)}     & 96758.2 (0.0012)     & 9340.4   &  & 96758 (0.0010)           & 96758.7 (0.0018)     & 6443.8   &  & \textbf{96757 (0.0000)}   & 96758.5 (0.0016)     & 8557.3   &  & 96780 (0.0238)             & 96789.5 (0.0336)     & 16446.8 \\
bm33708                   & 959289               & \textbf{959289 (0.0000)}    & 959291.4 (0.0003)    & 22647.9  &  & 959297 (0.0008)          & 959301.1 (0.0013)    & 16316.3  &  & 959291 (0.0002)           & 959297.2 (0.0009)    & 21874.8  &  & 959300 (0.0011)            & 959328.9 (0.0042)    & 16854.9 \\
pla33810                  & 66048945             & \textbf{66050069 (0.0017)}  & 66050888.1 (0.0029)  & 13383.5  &  & 66051574 (0.0040)        & 66054829.0 (0.0089)  & 6538.2   &  & 66053453 (0.0068)         & 66055157.3 (0.0094)  & 8817.9   &  & 66061997 (0.0198)          & 66071480.3 (0.0341)  & 16938.4 \\
pba38478                  & 108318               & \textbf{108318 (0.0000)}    & 108319.8 (0.0017)    & 15367.6  &  & 108319 (0.0009)          & 108321.1 (0.0029)    & 10023.5  &  & \textbf{108318 (0.0000)}  & 108319.6 (0.0015)    & 13529.7  &  & 108319 (0.0009)            & 108337.0 (0.0175)    & 19239.9 \\
ics39603                  & 106819               & 106820 (0.0009)             & 106822.2 (0.0030)    & 10214.3  &  & 106820 (0.0009)          & 106822.2 (0.0030)    & 8944.6   &  & \textbf{106819 (0.0000)}  & 106820.3 (0.0012)    & 11304.5  &  & 106823 (0.0037)            & 106834.2 (0.0142)    & 19802.2 \\
fht47608                  & 125104               & \textbf{125105 (0.0008)}    & 125107.5 (0.0028)    & 29345.7  &  & \textbf{125105 (0.0008)} & 125108.5 (0.0036)    & 20059.4  &  & \textbf{125105 (0.0008)}  & 125108.5 (0.0036)    & 26549.2  &  & 125128 (0.0192)            & 125143.7 (0.0317)    & 23805.0   \\
fna52057                  & 147789               & \textbf{147789 (0.0000)}    & 147790.7 (0.0012)    & 30942.0    &  & 147790 (0.0007)          & 147792.9 (0.0026)    & 22594.7  &  & 147790 (0.0007)           & 147792.9 (0.0026)    & 32481.8  &  & 147800 (0.0074)            & 147816.8 (0.0188)    & 26029.2 \\
bna56769                  & 158078               & \textbf{158078 (0.0000)}    & 158080.4 (0.0015)    & 36393.1  &  & 158079 (0.0006)          & 158080.6 (0.0016)    & 25839.3  &  & 158079 (0.0006)           & 158080.6 (0.0016)    & 35984.3  &  & 158105 (0.0171)            & 158117.7 (0.0251)    & 28384.8 \\
dan59296                  & 165371               & \textbf{165372 (0.0006)}    & 165373.3 (0.0014)    & 41479.0    &  & 165373 (0.0012)          & 165375.0 (0.0024)    & 30596.8  &  & 165373 (0.0012)           & 165375.0 (0.0024)    & 45269.8  &  & 165397 (0.0157)            & 165419.4 (0.0293)    & 29649.4 \\
ch71009                   & 4566506              & 4566508 (0.0000)            & 4566513.9 (0.0002)   & 205363.6 &  & 4566508 (0.0000)         & 4566518.0 (0.0003)   & 138965.6 &  & \textbf{4566507 (0.0000)} & 4566508.7 (0.0001)   & 194399.6 &  & 4566624 (0.0026)           & 4566887.5 (0.0084)   & 71013.5 \\
pla85900                  & 142382641            & \textbf{142384855 (0.0016)} & 142386004.9 (0.0024) & 70301.9  &  & 142390195 (0.0053)       & 142398897.0 (0.0114) & 64537.9  &  & 142388810 (0.0043)        & 142393735.8 (0.0078) & 85712.0    &  & 142409640 (0.0190)         & 142415681.4 (0.0232) & 85905.4 \\ \hline
Average                   &  --                    & \textbf{0.0004}            & \textbf{0.0026}     & 31513.5  &  & 0.0110                  & 0.0162              & 22697.5  &  & 0.0026                   & 0.0148              & 31515.0  &  & 0.0194                    & 0.0461              & 23877.9 \\ \bottomrule
\end{tabular}}
\end{sidewaystable}

\subsection{Comparing RHGA with EAX-GA and LKH}
We then present the detailed comparison of RHGA with EAX-300, EAX-400, and LKH, in solving all the 138 benchmark instances. The results on \textit{easy}, \textit{medium}, and \textit{hard} instances are shown in Tables \ref{table_Easy}, \ref{table_Medium}, and \ref{table_Hard}, respectively. We compare the best and average solutions in 10 runs obtained by the algorithms. We also provide the average calculation time of the algorithms. 

From the results in Tables \ref{table_Easy}, \ref{table_Medium}, and \ref{table_Hard}, we observe that:

(1) On all the 60 \textit{easy} instances, RHGA, EAX-300 and EAX-400 can easily yield the optimal solution in almost each of the 10 runs.

(2) On all the 62 \textit{medium} instances, RHGA exhibits better stability and robustness than EAX-300 and EAX-400, such as in solving the instances \textit{u1432}, \textit{fl1577}, \textit{pm8079}, and \textit{ar9152}. The average gap of the average solutions of RHGA is 84.5\% (85.1\%) less than that of EAX-300 (EAX-400).

(3) On all the 16 \textit{hard} instances, RHGA greatly outperforms EAX-300 and EAX-400. Specifically, the best solutions, the average solutions, and the worst solutions of RHGA are all better than those of EAX-300. The best solutions of RHGA are better than those of EAX-400 in 9 \textit{hard} instances, and worse than those of EAX-400 in 2 \textit{hard} instances. The average gap of the best solutions of RHGA is 96.4\% (84.6\%) less than that of EAX-300 (EAX-400), and the average gap of the average solutions of RHGA is 84.0\% (82.4\%) less than that of EAX-300 (EAX-400), indicating a significant improvement.

(4) The calculating time of RHGA and EAX-400 is close, indicating that RHGA can yield better performance than EAX-GA within the same parameters (compared to EAX-300) or similar calculation time (compared to EAX-400).

(5) The LKH is weaker than the other three algorithms in solving most of the tested instances. However, in solving the instances such as \textit{u2319} and \textit{pla7397}, LKH shows significantly better performance. Thus EAX-GA and LKH are complementary in solving different TSP instances, and our combination can make full use of their advantages and boost their performance.

\begin{figure*}[!t]
\centering
\subfigure[Cumulative gap on 51 \textit{small} but not \textit{easy} instances.]{
\includegraphics[width=0.96\columnwidth]{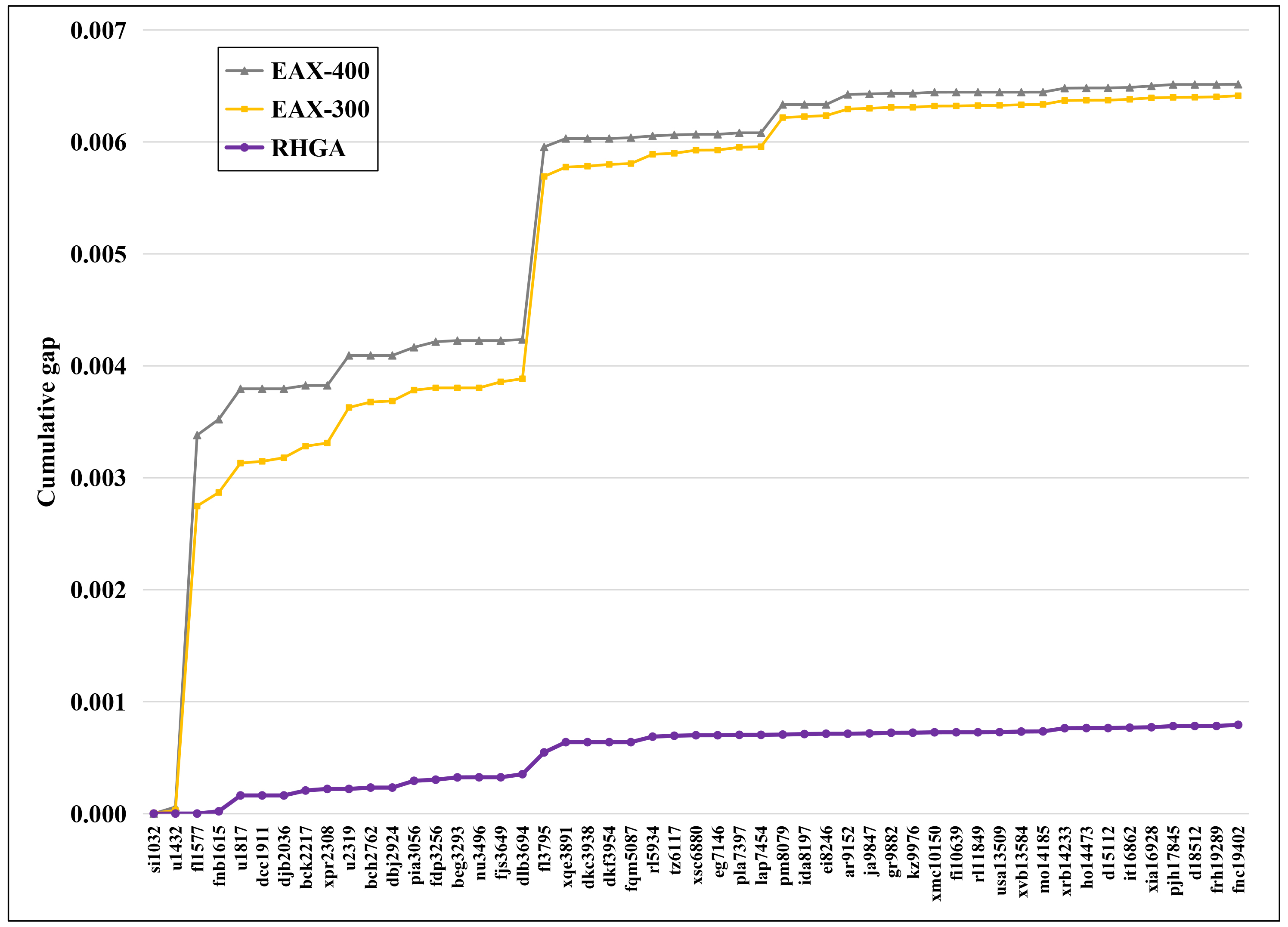} 
\label{fig_RHGA-EAX-small-a}
}
\subfigure[Cumulative run time on 51 \textit{small} but not \textit{easy} instances.]{
\includegraphics[width=0.96\columnwidth]{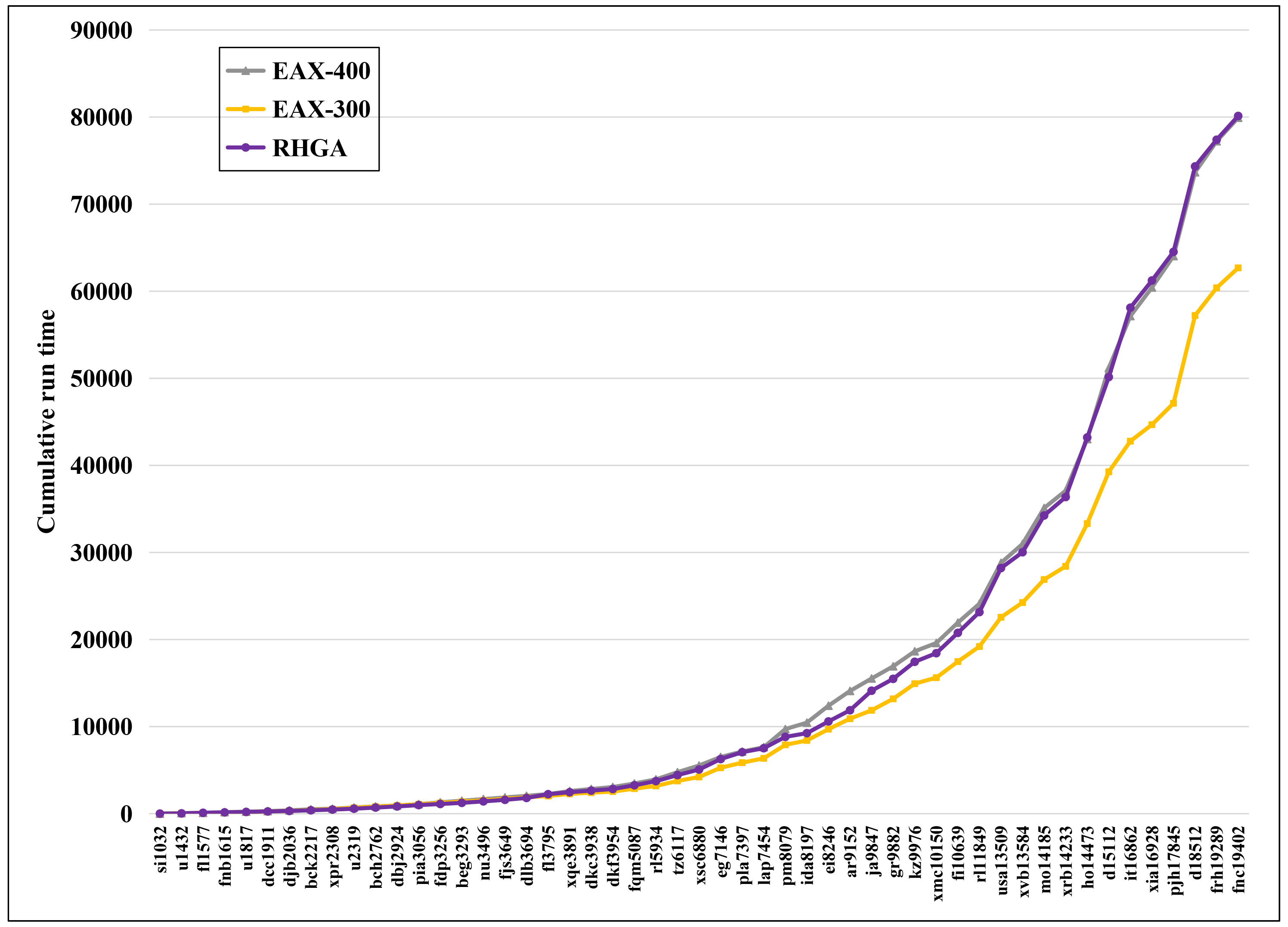} 
\label{fig_RHGA-EAX-small-b} 
}
\subfigure[Cumulative gap on 27 \textit{large} instances.]{
\includegraphics[width=0.96\columnwidth]{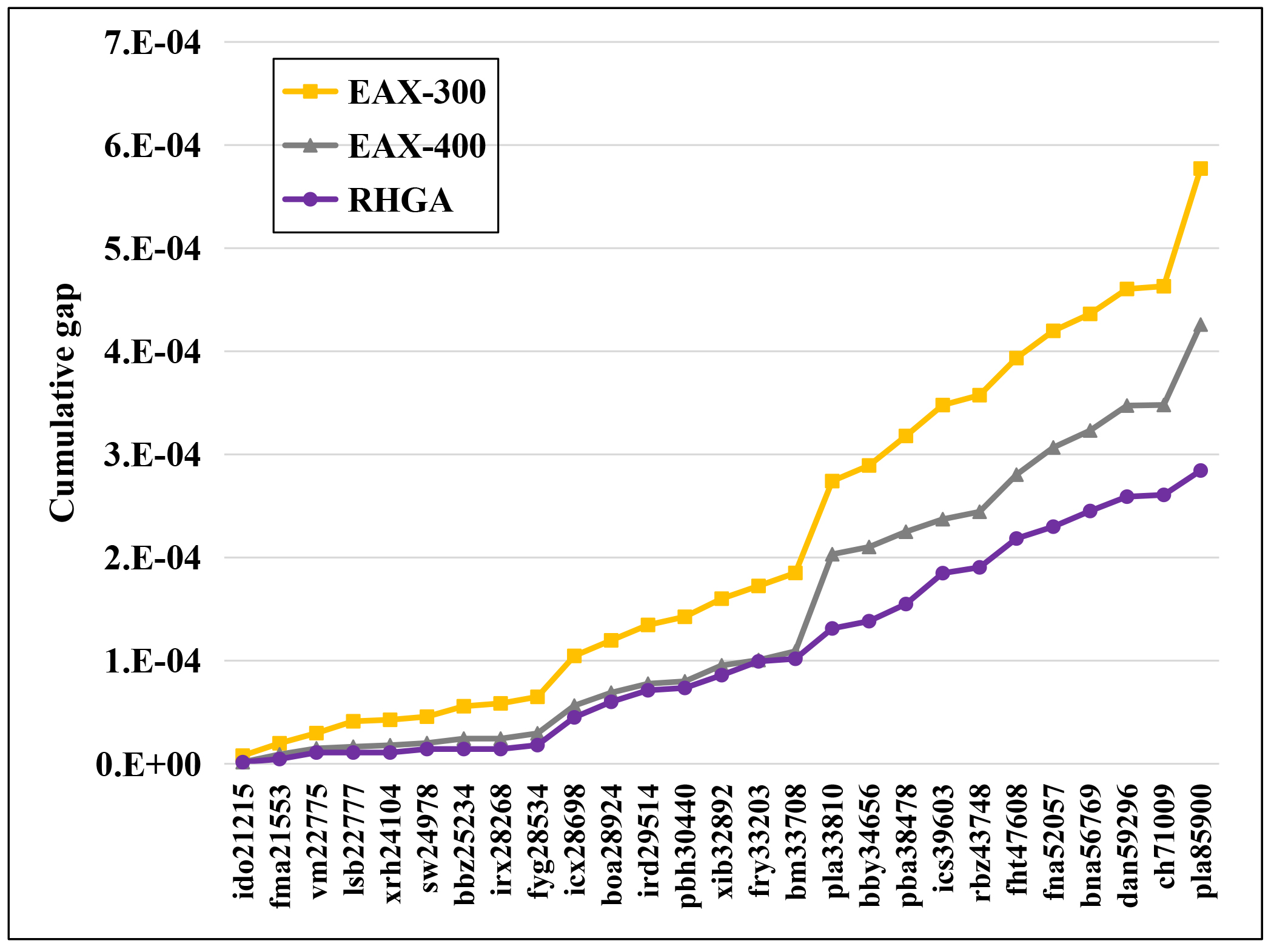} 
\label{fig_Large-a}
}
\subfigure[Cumulative run time on \textit{large} instances.]{
\includegraphics[width=0.96\columnwidth]{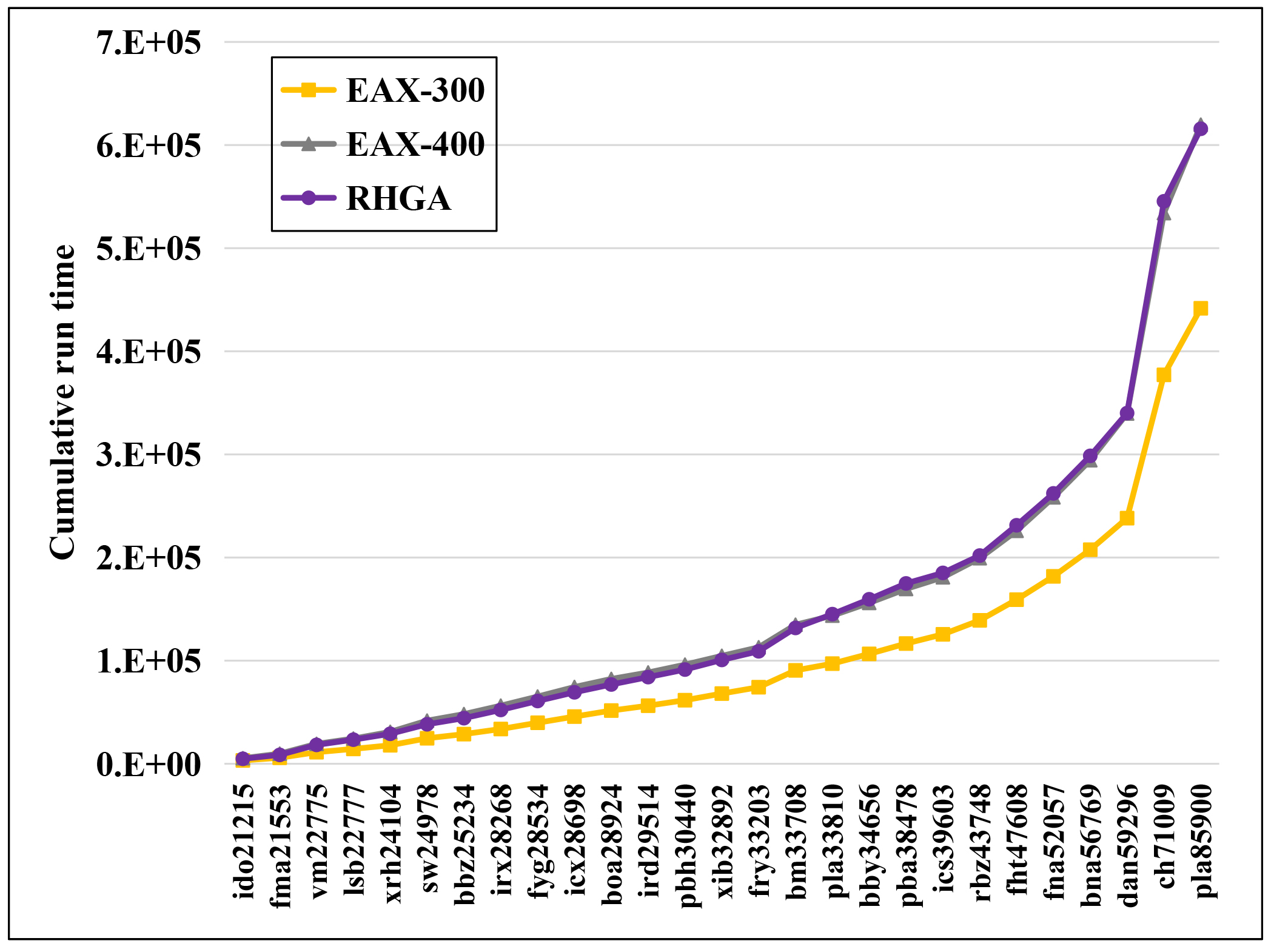} 
\label{fig_Large-b}
}
\caption{Comparison results of RHGA, EAX-300 and EAX-400 in solving the 51 \textit{small} but not \textit{easy} instances and 27 \textit{large} instances.}
\label{fig_RHGA-EAX}
\end{figure*}

To make a clearer comparison of RHGA and EAX-GA, we apply the cumulative metrics including cumulative gap on the solution quality and cumulative run time to compare RHGA with EAX-300 and EAX-400 in solving all the 51 \textit{small} but not \textit{easy} instances and all the 27 \textit{large} instances. Let $gap(j)=\frac{1}{10}\sum_{i=1}^{10}\frac{A_i-BKS_j}{BKS_j}$ be the average gap of calculating the $j$-th instance by an algorithm in 10 runs, where $A_i$ is the result of the $i$-th calculation and $BKS_j$ is the best-known solution of the $j$-th instance. The smaller the average gap, the closer the average solution is to the best-known solution. For an algorithm, $C_{gap}(j)=\sum_{i=1}^{j}gap(i)$ is the cumulative gap. The cumulative run time can be calculated similarly. The comparison results are shown in Figure \ref{fig_RHGA-EAX}.

The results indicate again that the robustness and stability of EAX-GA are not good. It shows much worse performance than RHGA in solving some instances, such as \textit{fl1577}, \textit{u2319}, \textit{fl3795}, \textit{pla33810}, and \textit{pla85900}. The increase of population size from 300 to 400 can not help EAX-GA escape from the local optima when solving these instances, while the proposed methods including the hybrid mechanism and reinforcement learning could. We can also observe that the calculation time of RHGA is close to EAX-400 and a little bit longer than that of EAX-300. The results clearly show again that RHGA significantly outperforms EAX-GA within the same parameters or similar calculation time.

\section{Further Analysis}
\label{Sec_Analysis}
This section provides insight on why and how the proposed RHGA is effective, suggesting that the creative combination of EAX-GA and LKH can boost the performance mutually. The results further indicate that the adaptive Q-value learned by reinforcement learning is a better metric for determining the candidate cities and evaluating the edge quality than the $\alpha$-value in LKH as well as the distance in EAX-GA. We first introduce various variant algorithms involved in the experiments, then present and analyze the results.

\subsection{Various Variants of the Algorithm}
We first present various variant algorithms of RHGA and EAX-GA for comparison and analysis. The variant algorithms include the following:

\begin{itemize}
\item \textbf{Alpha-EAX}: A variant of EAX-300 using the $\alpha$-value (Eq. \ref{eq_alpha-value}) to replace the distance metric used in the process of merging sub-tours in EAX-300 (see Step 5 in Section \ref{sec_EAX}).
\item \textbf{FixQ-EAX}: A variant of EAX-300 using the initial Q-value (Eq. \ref{eq_InitQ}) to replace the distance metric used in the process of merging sub-tours in EAX-300.
\item \textbf{Q-EAX}: A variant of EAX-300 using the adaptive Q-value learned by the Q-learning method (Eq. \ref{eq_QLearning}) to replace the distance metric used in the process of merging sub-tours in EAX-300. The population size in Q-EAX is set to be 301. The extra individual is the special individual used to learn the Q-value and provide the adaptive Q-value for the algorithm. The special individual in Q-EAX can be improved by the Q-LKH local search and the genetic algorithm, but other individuals cannot be improved by crossing with the special individual (i.e., the special individual in Q-EAX can only be $p_A$, not $p_B$). In addition, the result of Q-EAX is the best individual in the population except for the special individual, and Q-EAX will not terminate if the special individual is the optimum solution (if known).
\item \textbf{Q-EAX+Special}: A variant of Q-EAX that the output is the best individual in the population includes the special individual. Q-EAX+Special terminates when the special individual is the optimum solution. In a word, the difference between Q-EAX+Special and Q-EAX includes: 1) whether the special individual can be the output solution. 2) Whether the algorithm terminates when the special individual is the optimum solution.
\item \textbf{EAX-LKH}: A variant of RHGA that combines EAX-GA with the LKH algorithm as RHGA does but with no reinforcement learning. The metrics in the EAX-GA and LKH local search are not changed.
\item \textbf{Alpha-EAX-LKH}: A variant of EAX-LKH using the $\alpha$-value (Eq. \ref{eq_alpha-value}) to replace the distance metric in EAX-GA.
\item \textbf{FixQ-EAX-LKH}: A variant of EAX-LKH using the initial Q-value (Eq. \ref{eq_InitQ}) to replace the distance metric in EAX-GA and the $\alpha$-value metric in LKH.
\item \textbf{RHGA-$k$}: A variant of RHGA that the number of the special individuals is $k$ (we tested $k=3, 5, 20, 50$ in experiments). Each special individual $x_i$, $i\leq{k}$, can be improved by Q-LKH when: 1) it is just initialized, or it was improved by EAX-GA at the last generation. In this case, Q-LKH will try to improve $x_i$. 2) It has not been improved for $M_{gen}$ generations, and the algorithm randomly selects an individual $x_r$ ($x_r$ is not a special individual). If $x_r$ can be improved by Q-LKH and the improved tour is better than $x_i$, then replace $x_i$ with the improved tour. Moreover, when the best individual $x_{best}$ in the population besides the $k$ special individuals is better than the best special individual and $x_{best}$ has not been calculated by Q-LKH, Q-LKH will try to improve $x_{best}$. The improved tour will replace $x_1$ (do not replace $x_1$ if $x_{best}$ cannot be improved by Q-LKH).
\end{itemize}

The termination conditions of all the above algorithms are the same as in RHGA. Since the calculation time of RHGA is a little longer than EAX-300 and close to EAX-400 when solving the tested benchmarks, the calculation times of the variant algorithms, including Alpha-EAX, FixQ-EAX, Q-EAX, Q-EAX+Special, EAX-LKH, Alpha-EAX-LKH, FixQ-EAX-LKH, are also roughly between that of EAX-300 and EAX-400.

We further introduce three hybrid algorithms that combine EAX-GA with LKH in a straightforward manner:

\begin{itemize}
\item \textbf{Min\{EAX, LKH\}}: Given an input instance, this algorithm first uses EAX-GA and LKH to calculate the instance 10 times respectively, and then selects the one with the better average solution as the output.
\item \textbf{LKH+EAX}: A hybrid algorithm that uses LKH to generate the initial population, and then runs EAX-GA starting from this population.
\item \textbf{EAX+LKH}: A hybrid algorithm that first uses EAX-GA to calculate the input instance until the termination conditions of EAX-GA are reached (or the known optimum is obtained), then runs LKH with its initial solution equal to the best individual. The LKH algorithm in EAX+LKH terminates as EAX-GA does. That is, let $Gen$ be the number of iterations at which no improvement in the best solution is found by LKH over the recent $1500/N_{ch}$ iterations. If $Gen$ has been determined and the best solution does not improve over the last $G_{max} = Gen/10$ iterations, EAX+LKH terminates. 
\end{itemize}

\subsection{Analysis on the Combination Mechanism of RHGA}
\begin{figure*}[!t]
\centering
\subfigure[Comparison results with LKH+EAX.]{
\includegraphics[width=0.96\columnwidth]{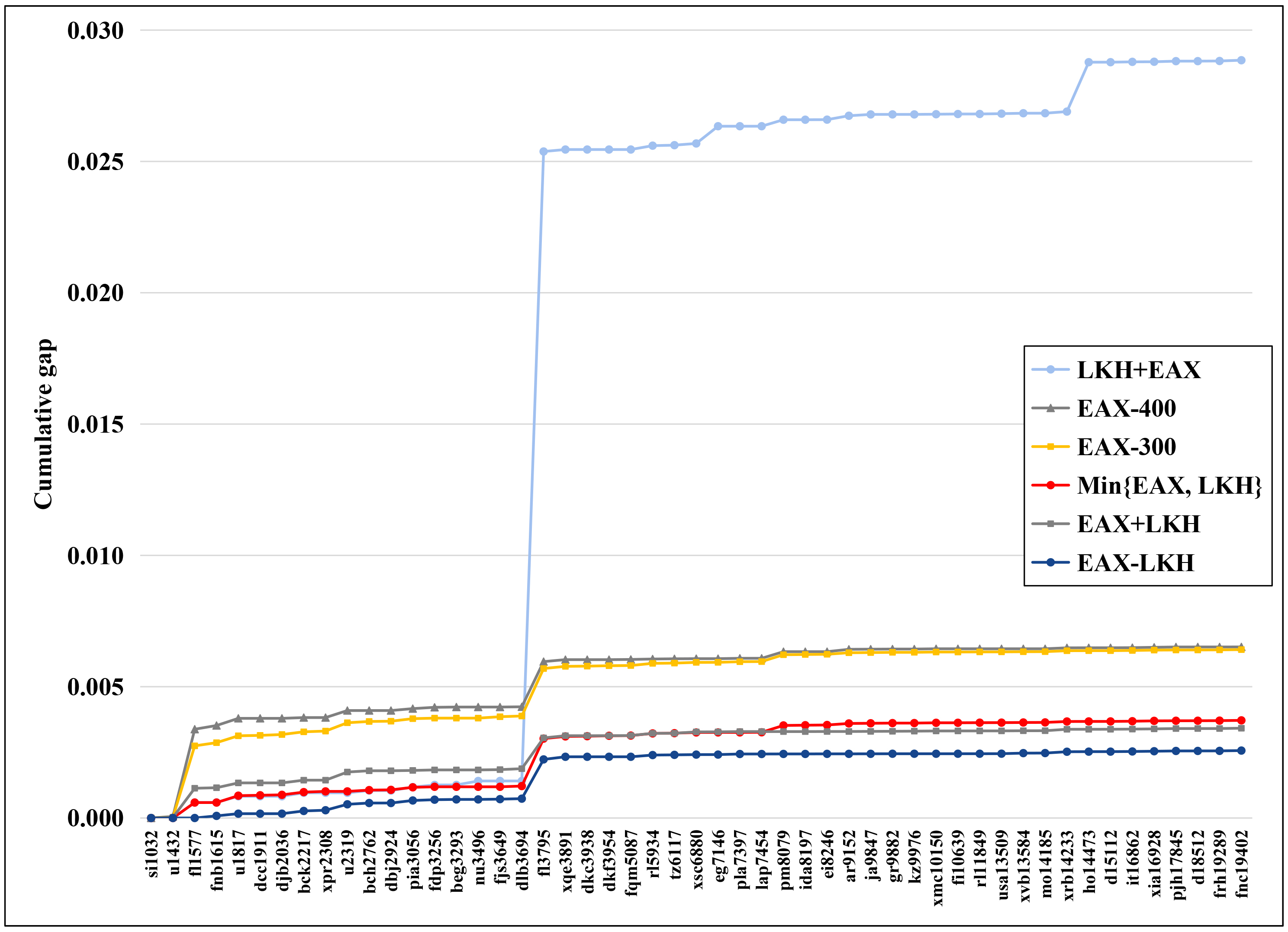} 
\label{fig_Hybrid-a}
}
\subfigure[Comparison results without LKH+EAX.]{
\includegraphics[width=0.96\columnwidth]{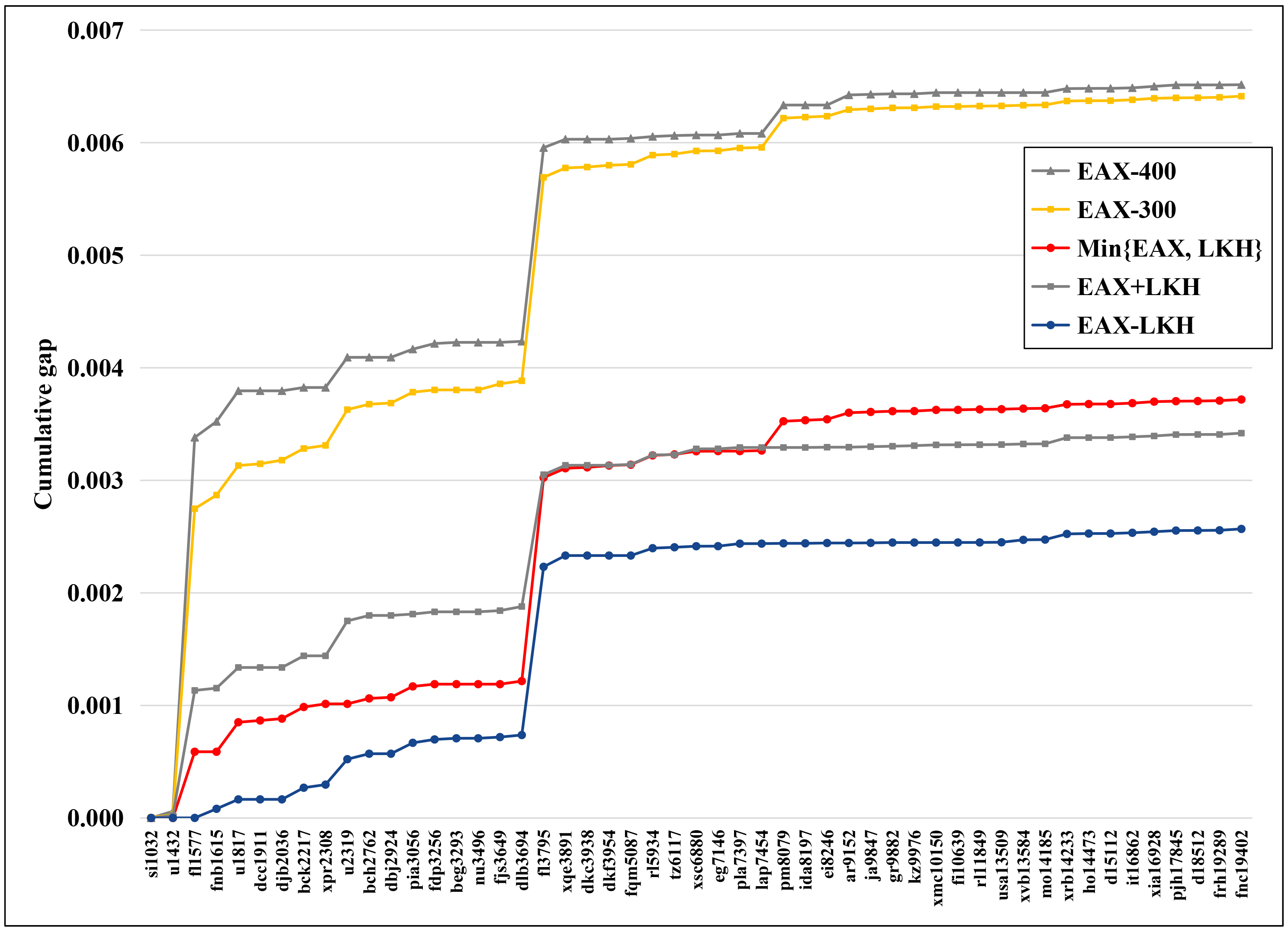} 
\label{fig_Hybrid-b} 
}
\caption{Analysis on the combination mechanism on 51 \textit{small} but not \textit{easy} instances.}
\label{fig_Hybrid}
\end{figure*}

In order to demonstrate that our proposed combination mechanism is reasonable and effective, we first compare EAX-LKH with the hybrid algorithms Min\{EAX, LKH\}, LKH+EAX, and EAX+LKH, as well as the baselines EAX-300 and EAX-400 on all the 51 \textit{small} but not \textit{easy} instances in Figure \ref{fig_Hybrid-a}. We also present the results without LKH+EAX in Figure \ref{fig_Hybrid-b} for a clearer comparison. 

As shown in Figure \ref{fig_Hybrid}, LKH+EAX is much worse than EAX-300 and EAX-400. Note that the method to generate the initial population in EAX-GA is a simple greedy 2-opt local search~\cite{Nagata2013}. Why does LKH+EAX use the effective LKH local search to replace the simple 2-opt results in much worse performance? The reason is that the initial population of LKH+EAX contains too many candidate edges provided by LKH. Thus the population diversity is broken and it is easy for the algorithm to get stuck in local optima. In this case, if LKH can provide good genes, i.e., the edges in the optimal solution are contained in the candidate edges, LKH+EAX can obtain better results than EAX-GA. Otherwise, its performance is poor. The results in Figure \ref{fig_Hybrid} can demonstrate this comment, as the performance of LKH+EAX mainly depends on LKH. For the instances that LKH works well (see detailed results of LKH in Tables \ref{table_Easy}, \ref{table_Medium}, and \ref{table_Hard}), such as \textit{u1432} and \textit{u2319}, LKH+EAX shows better performance than EAX-300 and EAX-400. For the instances that LKH can not work well, such as \textit{fl3795}, \textit{eg7146}, and \textit{ho14472}. LKH+EAX shows much worse performance than EAX-300 and EAX-400. This result can also demonstrate that our mechanism that only one special individual can be improved by LKH is reasonable. 

Moreover, algorithms Min\{EAX, LKH\} and EAX+LKH can surely obtain results no worse than EAX-GA, because they can obtain at least the same solution as EAX-GA does. However, the straightforward combinations of EAX-GA and LKH can not make full use of their complementary, as the EAX-LKH algorithm with our combination mechanism shows much better performance than Min\{EAX, LKH\} and EAX+LKH. Note that Min\{EAX, LKH\} can be regarded as the hybrid solver proposed by Kerschke et al.~\cite{Kerschke2018} consists of EAX-GA and LKH with a perfect prediction model. Therefore, the combination mechanism in RHGA is much more effective than the straightforward hybrid mechanisms we designed for comparison and the hybrid mechanism proposed by Kerschke et al.~\cite{Kerschke2018}, and can make better use of the complementary of EAX-GA and LKH.

\subsection{Analysis on Different Metrics and the Special Individual}
\begin{figure*}[!t]
\centering
\subfigure[Comparison results with EAX-300 and EAX-400.]{
\includegraphics[width=0.96\columnwidth]{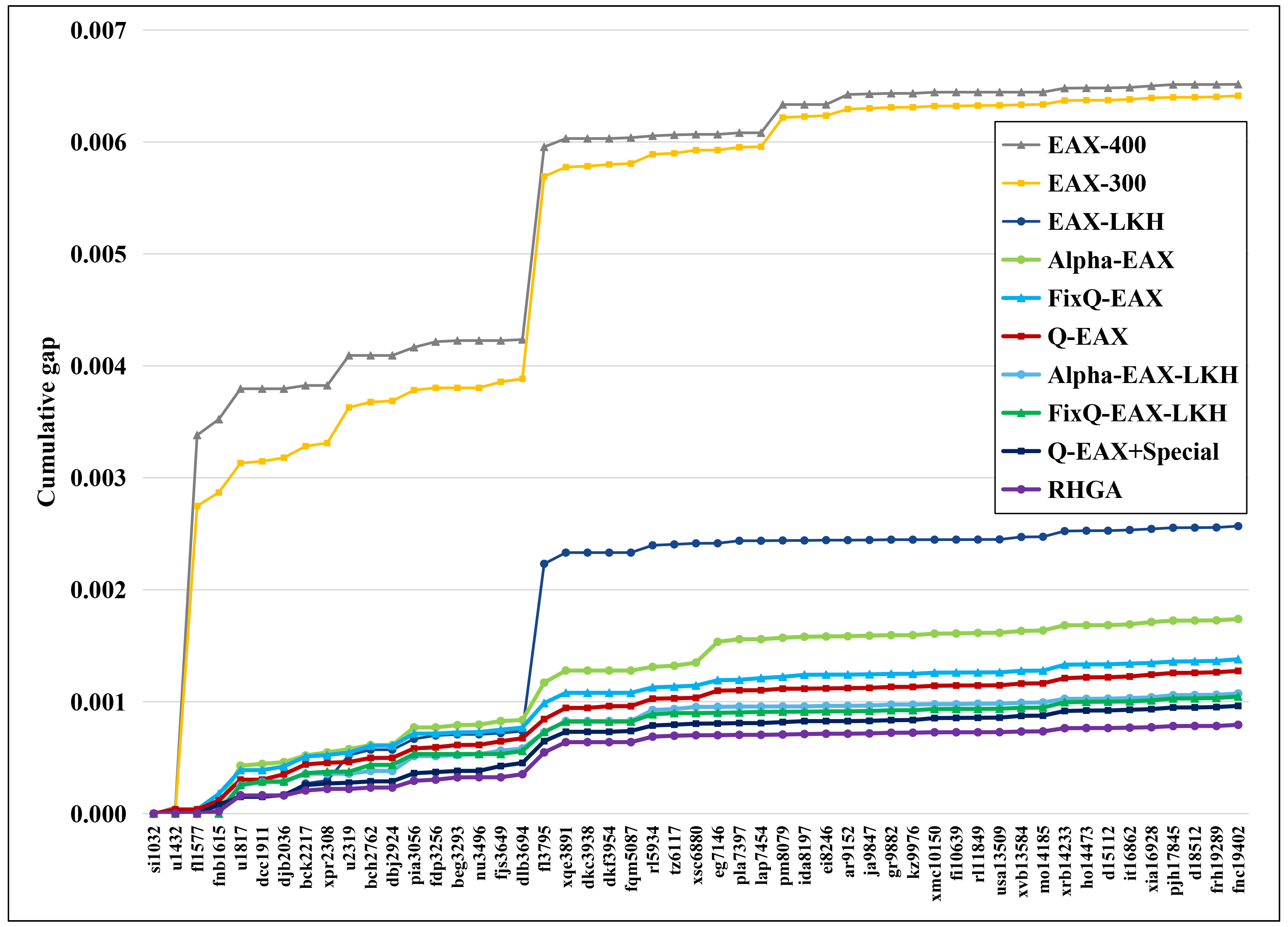} 
\label{fig_SmallAll-a}
}
\subfigure[Comparison results without EAX-300 and EAX-400.]{
\includegraphics[width=0.96\columnwidth]{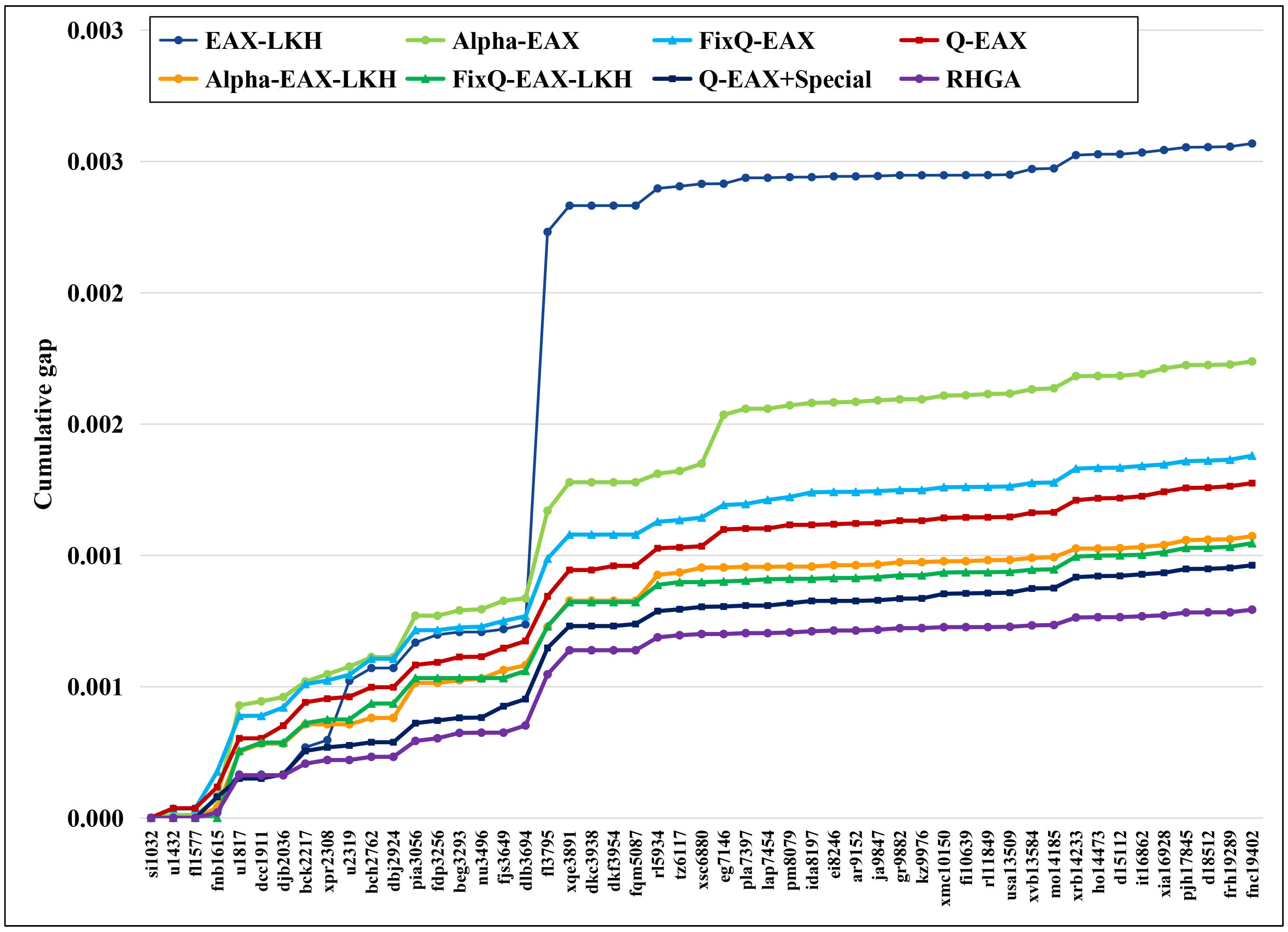} 
\label{fig_SmallAll-b} 
}
\caption{Analysis on different metrics and the special individual on 51 \textit{small} but not \textit{easy} instances.}
\label{fig_SmallAll}
\end{figure*}

We then compare RHGA with the variant algorithms to evaluate the performance of different metrics in determining the candidate edges, and evaluate the effectiveness of the special individual in RHGA. Figure \ref{fig_SmallAll-a} compares the results of 10 different algorithms, including RHGA, Alpha-EAX, FixQ-EAX, Q-EAX, Q-EAX+Special, EAX-LKH, Alpha-EAX-LKH, FixQ-EAX-LKH, EAX-300, and EAX-400, on the same 51 instances in Figure \ref{fig_Hybrid}. We also present the results without EAX-300 and EAX-400 in Figure \ref{fig_SmallAll-b} for a clearer comparison. 

\begin{figure*}[!t]
\centering
\subfigure[Comparison on cumulative gap.]{
\includegraphics[width=0.96\columnwidth]{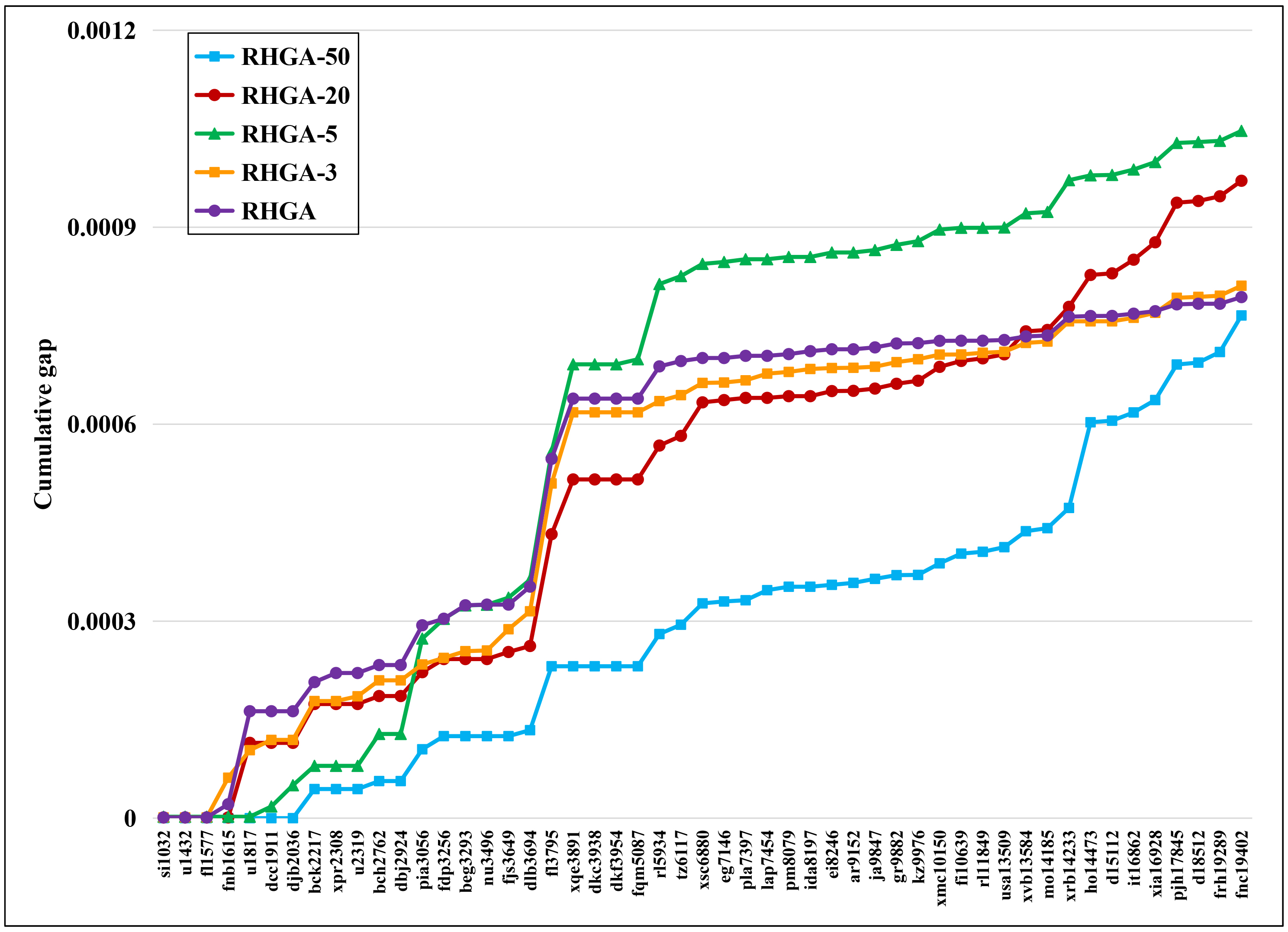} 
\label{fig_RHGA-k-a}
}
\subfigure[Comparison on cumulative run time.]{
\includegraphics[width=0.96\columnwidth]{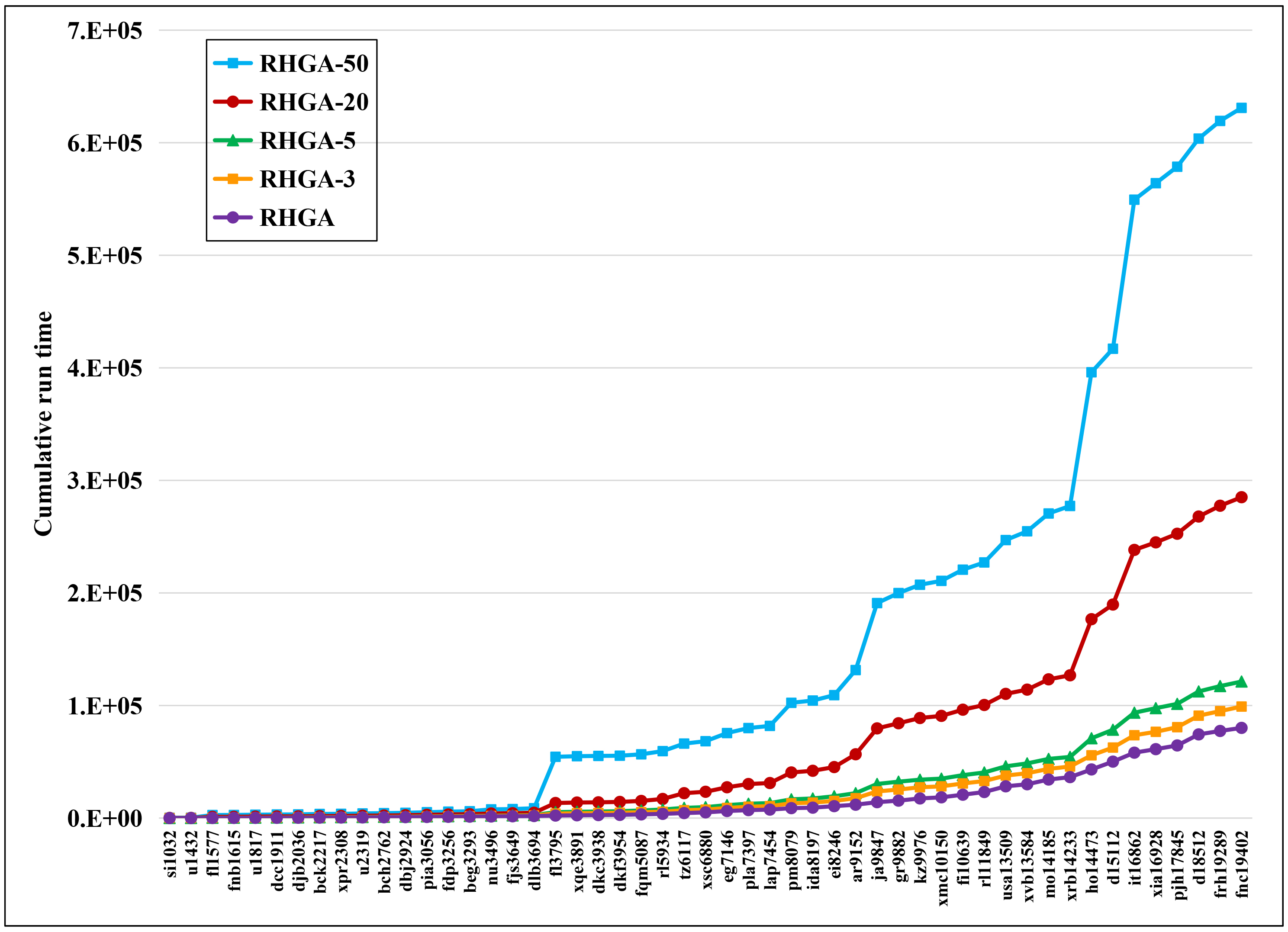} 
\label{fig_RHGA-k-b} 
}
\caption{Comparison results of RHGA and RHGA-3/5/20/50 in solving the 51 \textit{small} but not \textit{easy} instances.\vspace{-1em}}
\label{fig_RHGA-k}
\end{figure*}

The results of EAX-LKH, Alpha-EAX-LKH, FixQ-EAX-LKH, and RHGA in Figure \ref{fig_SmallAll} indicate that, the order with decaying quality of the performance of different metrics is: adaptive Q-value (updated by reinforcement learning according to Eq. \ref{eq_QLearning}), initial Q-value (Eq. \ref{eq_InitQ}), $\alpha$-value and distance. The results of EAX-300, Alpha-EAX, FixQ-EAX, and Q-EAX can draw the same conclusion. As a result, the EAX-GA algorithm can be improved simply by replacing the distance metric with the $\alpha$-value or initial Q-value, and reinforcement learning can further improve the performance of the initial Q-value. Moreover, the combination of EAX-GA and LKH by our proposed mechanism is always effective, since EAX-LKH/Alpha-EAX-LKH/FixQ-EAX-LKH/RHGA outperforms EAX-300/Alpha-EAX/FixQ-EAX/Q-EAX significantly.

The results of Q-EAX, Q-EAX+Special, and RHGA in Figure \ref{fig_SmallAll} can further lead to the following comments. First, Q-EAX+Special outperforms Q-EAX, indicating that the special individual can provide better solutions than the other individuals. Second, RHGA outperforms Q-EAX+Special, indicating that the special individual can also improve the population by spreading its genes to other individuals. In summary, the benefits of the special individual are as follows: 1) it can provide the adaptive Q-value for the EAX-GA to improve the performance. 2) it can obtain good solutions since it can be improved by both Q-LKH and EAX-GA (when $x_1$ is parent $p_A$). 3) it can improve the population by providing its high-quality genes for other individuals (when $x_1$ is parent $p_B$).

\subsection{Comparison with Multiple Special Individuals}
\label{sec_indi}

We then compare RHGA with the RHGA-$k$ algorithms, including RHGA-3, RHGA-5, RHGA-20, and RHGA-50, on the same 51 instances in Figure \ref{fig_Hybrid}, to analyze the influence of the number of special individuals on the performance. The comparison results are shown in Figure \ref{fig_RHGA-k}.

From the results 
we can observe that:

(1) RHGA and RHGA-3 show similar performance in solving the 51 instances. The algorithms with a larger number of special individuals show better performance in solving the instances with relatively small scales. For example, the RHGA-5 is good at solving the instances with less than 3,000 cities, and the RHGA-20 and RHGA-50 are good at solving the instances with less than 6,000 cities. However, 
neither of the three algorithms with a larger number of special individuals is good at solving the instances with larger than 14,000 cities,
indicating that for solving relatively large and complex instances, an excessive proportion of special individuals in the population may reduce the algorithm performance.

(2) The calculation time of the RHGA-$k$ algorithm increases rapidly as $k$ increases. Specifically, the calculation time of RHGA-3/5/20/50 is about 24/51/256/687\% more than that of RHGA.

In summary, setting multiple special individuals in the population is not reasonable 
and time-consuming. Thus it is suitable and effective to set only one special individual in the population, which can help the algorithm obtain a good performance without reducing the efficiency.

\section{Conclusion}
\label{Sec_Con}
In this paper, we address the famous NP-hard traveling salesman problem and propose a reinforced hybrid genetic algorithm (RHGA) that combines reinforcement learning with two state-of-the-art heuristic algorithms, the EAX-GA genetic algorithm and the LKH local search heuristic, in a more interactive form. The EAX-GA and LKH are integrated with the help of a special and unique individual, which can be improved by both the genetic algorithm and the local search algorithm. In our proposed hybrid mechanism, the population provides diverse and high-quality initial solutions for the LKH local search algorithm, and the local search algorithm leads the population to converge to better results. In a word, the two state-of-the-art TSP heuristics, EAX-GA and LKH, can boost each other in our proposed hybrid mechanism. 
Moreover, the Q-learning algorithm is applied to learn an adaptive Q-value to replace the distance metric used in the process of merging sub-tours in EAX-GA and the $\alpha$-value used in LKH for determining the candidate cities. As a result, our reinforcement learning method can improve both the EAX-GA and the LKH algorithms by providing better candidate edges.

Extensive experimental results demonstrate that RHGA outperforms the powerful EAX-GA and LKH algorithms, as well as one of the state-of-the-art (deep) learning based algorithms, NeuroLKH, for solving the TSP. Further and extensive ablation studies are adopted to show the effectiveness of the proposed hybrid mechanism and the reinforcement learning method, and to demonstrate that the proposed hybrid mechanism can make full use of the complementary of EAX-GA and LKH, and the setting of only one special individual is reasonable and efficient. 

In future work, the proposed mechanism of combining genetic algorithms with local search could be applied to solve various combinatorial optimization problems, and the method of combining reinforcement learning with the core process of heuristics would also be applied to improve other heuristic algorithms.


\bibliographystyle{unsrt}

\bibliography{RHGA}

\begin{thebibliography}{10}

\bibitem{Khalil2017}
Elias~B. Khalil, Hanjun Dai, Yuyu Zhang, Bistra Dilkina, and Le~Song.
\newblock Learning combinatorial optimization algorithms over graphs.
\newblock In {\em Annual Conference on Neural Information Processing Systems
  2017, NIPS 2017}, pages 6348--6358, 2017.

\bibitem{Helsgaun2000}
Keld Helsgaun.
\newblock An effective implementation of the lin-kernighan traveling salesman
  heuristic.
\newblock {\em European Journal of Operational Research}, 126(1):106--130,
  2000.

\bibitem{Nagata2013}
Yuichi Nagata and Shigenobu Kobayashi.
\newblock A powerful genetic algorithm using edge assembly crossover for the
  traveling salesman problem.
\newblock {\em {INFORMS} Journal on Computing}, 25(2):346--363, 2013.

\bibitem{Lin1973}
Shen Lin and Brian~W. Kernighan.
\newblock An effective heuristic algorithm for the traveling-salesman problem.
\newblock {\em Operations Research}, 21(2):498--516, 1973.

\bibitem{Tsai2004}
Huai{-}Kuang Tsai, Jinn{-}Moon Yang, Yuan{-}Fang Tsai, and Cheng{-}Yan Kao.
\newblock An evolutionary algorithm for large traveling salesman problems.
\newblock {\em {IEEE} Transactions on Systems, Man, and Cybernetics, Part B},
  34(4):1718--1729, 2004.

\bibitem{Nagata1997}
Yuichi Nagata and Shigenobu Kobayashi.
\newblock Edge assembly crossover: A high-power genetic algorithm for the
  travelling salesman problem.
\newblock In {\em The Seventh International Conference on Genetic Algorithms,
  ICGA 1997}, pages 450--457, 1997.

\bibitem{Ulder1990}
Nico~LJ Ulder, Emile~HL Aarts, Hans-J{\"u}rgen Bandelt, Peter JM~Van Laarhoven,
  and Erwin Pesch.
\newblock Genetic local search algorithms for the traveling salesman problem.
\newblock In {\em International Conference on Parallel Problem Solving from
  Nature, PPSN 1990}, volume 496, pages 109--116, 1990.

\bibitem{Freisleben1996}
Bernd Freisleben and Peter Merz.
\newblock New genetic local search operators for the traveling salesman
  problem.
\newblock In {\em Parallel Problem Solving from Nature - {PPSN} IV,
  International Conference on Evolutionary Computation. The Fourth
  International Conference on Parallel Problem Solving from Nature, PPSN 1996},
  volume 1141, pages 890--899, 1996.

\bibitem{Merz1997}
Peter Merz and Bernd Freisleben.
\newblock Genetic local search for the tsp: new results.
\newblock In {\em {IEEE} International Conference on Evolutionary Computation,
  ICEC 1997}, pages 159--164, 1997.

\bibitem{Nguyen2007}
Hung~Dinh Nguyen, Ikuo Yoshihara, Kunihito Yamamori, and Moritoshi Yasunaga.
\newblock Implementation of an effective hybrid {GA} for large-scale traveling
  salesman problems.
\newblock {\em {IEEE} Transactions on Systems, Man, and Cybernetics, Part B},
  37(1):92--99, 2007.

\bibitem{Whitley2010}
L.~Darrell Whitley, Doug Hains, and Adele~E. Howe.
\newblock A hybrid genetic algorithm for the traveling salesman problem using
  generalized partition crossover.
\newblock In {\em Parallel Problem Solving from Nature - {PPSN} XI, Eleventh
  International Conference, Part {I}, PPSN {I} 2010}, volume 6238, pages
  566--575, 2010.

\bibitem{Wang2014}
Yong Wang.
\newblock The hybrid genetic algorithm with two local optimization strategies
  for traveling salesman problem.
\newblock {\em Computers {\&} Industrial Engineering}, 70:124--133, 2014.

\bibitem{Rashid2017}
Mohammad~Harun Rashid and Miguel~A. Mosteiro.
\newblock A greedy-genetic local-search heuristic for the traveling salesman
  problem.
\newblock In {\em 2017 {IEEE} International Symposium on Parallel and
  Distributed Processing with Applications and 2017 {IEEE} International
  Conference on Ubiquitous Computing and Communications, ISPA/IUCC 2017}, pages
  868--872, 2017.

\bibitem{Ilin2020}
Vladimir Ilin, Dragan Simic, Svetislav~D. Simic, and Svetlana Simic.
\newblock Hybrid genetic algorithms and tour construction and improvement
  algorithms used for optimizing the traveling salesman problem.
\newblock In {\em Fifteenth International Conference on Soft Computing Models
  in Industrial and Environmental Applications, {SOCO} 2020}, volume 1268,
  pages 530--539, 2020.

\bibitem{Kerschke2018}
Pascal Kerschke, Lars Kotthoff, Jakob Bossek, Holger~H. Hoos, and Heike
  Trautmann.
\newblock Leveraging {TSP} solver complementarity through machine learning.
\newblock {\em Evolutionary Computation}, 26(4), 2018.

\bibitem{Sutton1998}
Richard~S. Sutton and Andrew~G. Barto.
\newblock {\em Reinforcement learning - an introduction}.
\newblock Adaptive computation and machine learning. {MIT} Press, 1998.

\bibitem{Hu2020}
Yujiao Hu, Yuan Yao, and Wee~Sun Lee.
\newblock A reinforcement learning approach for optimizing multiple traveling
  salesman problems over graphs.
\newblock {\em Knowledge-Based Systems}, 204:106244, 2020.

\bibitem{Oladayo2022}
Ajani~S. Oladayo and Rammohan Mallipeddi.
\newblock Adaptive evolution strategy with ensemble of mutations for
  reinforcement learning.
\newblock {\em Knowledge-Based Systems}, 245:108624, 2022.

\bibitem{Zheng2021}
Jiongzhi Zheng, Kun He, Jianrong Zhou, Yan Jin, and Chu{-}min Li.
\newblock Combining reinforcement learning with lin-kernighan-helsgaun
  algorithm for the traveling salesman problem.
\newblock In {\em Thirty-Fifth {AAAI} Conference on Artificial Intelligence,
  {AAAI} 2021}, pages 12445--12452, 2021.

\bibitem{Bengio2021}
Yoshua Bengio, Andrea Lodi, and Antoine Prouvost.
\newblock Machine learning for combinatorial optimization: {A} methodological
  tour d'horizon.
\newblock {\em European Journal of Operational Research}, 290(2):405--421,
  2021.

\bibitem{Bello2017}
Irwan Bello, Hieu Pham, Quoc~V. Le, Mohammad Norouzi, and Samy Bengio.
\newblock Neural combinatorial optimization with reinforcement learning.
\newblock In {\em Fitth International Conference on Learning Representations,
  Workshop Track Proceedings, {ICLR} (Workshop) 2017}, 2017.

\bibitem{Vinyals2015}
Oriol Vinyals, Meire Fortunato, and Navdeep Jaitly.
\newblock Pointer networks.
\newblock In {\em Annual Conference on Neural Information Processing Systems
  2015, NIPS 2015}, pages 2692--2700, 2015.

\bibitem{Goh2022}
Yong~Liang Goh, Wee~Sun Lee, Xavier Bresson, Thomas Laurent, and Nicholas Lim.
\newblock Combining reinforcement learning and optimal transport for the
  traveling salesman problem.
\newblock {\em CoRR}, abs/2203.00903, 2022.

\bibitem{Cuturi2013}
Marco Cuturi.
\newblock Sinkhorn distances: Lightspeed computation of optimal transport.
\newblock In {\em Annual Conference on Neural Information Processing Systems
  2013.}, pages 2292--2300, 2013.

\bibitem{Emami2018}
Patrick Emami and Sanjay Ranka.
\newblock Learning permutations with sinkhorn policy gradient.
\newblock {\em CoRR}, abs/1805.07010, 2018.

\bibitem{Liu2009}
Fei Liu and Guangzhou Zeng.
\newblock Study of genetic algorithm with reinforcement learning to solve the
  {TSP}.
\newblock {\em Expert Systems with Applications}, 36(3):6995--7001, 2009.

\bibitem{Nagata2006}
Yuichi Nagata.
\newblock New {EAX} crossover for large {TSP} instances.
\newblock In {\em Parallel Problem Solving from Nature - {PPSN} IX, Ninth
  International Conference, PPSN 2006}, volume 4193, pages 372--381, 2006.

\bibitem{Costa2020}
Paulo~R. de~O.~da Costa, Jason Rhuggenaath, Yingqian Zhang, and Alp Akcay.
\newblock Learning 2-opt heuristics for the traveling salesman problem via deep
  reinforcement learning.
\newblock In {\em The Twelfth Asian Conference on Machine Learning, ACML 2020},
  volume 129, pages 465--480, 2020.

\bibitem{Sui2021}
Jingyan Sui, Shizhe Ding, Ruizhi Liu, Liming Xu, and Dongbo Bu.
\newblock Learning 3-opt heuristics for traveling salesman problem via deep
  reinforcement learning.
\newblock In {\em The Thirteenth Asian Conference on Machine Learning, ACML
  2021}, volume 157, pages 1301--1316, 2021.

\bibitem{Zhao2021}
Jiuxia Zhao, Minjia Mao, Xi~Zhao, and Jianhua Zou.
\newblock A hybrid of deep reinforcement learning and local search for the
  vehicle routing problems.
\newblock {\em {IEEE} Transactions on Intelligent Transportation Systems},
  22(11):7208--7218, 2021.

\bibitem{Xin2021}
Liang Xin, Wen Song, Zhiguang Cao, and Jie Zhang.
\newblock Neurolkh: Combining deep learning model with lin-kernighan-helsgaun
  heuristic for solving the traveling salesman problem.
\newblock In {\em Annual Conference on Neural Information Processing Systems
  2021, NeurIPS 2021}, pages 7472--7483, 2021.

\bibitem{Maekawa1996}
Keiji Maekawa, Naoki Mori, Hisashi Tamaki, Hajime Kita, and Yoshikazu
  Nishikawa.
\newblock A genetic solution for the traveling salesman problem by means of a
  thermodynamical selection rule.
\newblock In {\em {IEEE} International Conference on Evolutionary Computation,
  ICEC 1996}, pages 529--534, 1996.

\bibitem{Lin1965}
Shen Lin.
\newblock Computer solutions of the traveling salesman problem.
\newblock {\em Bell System Technical Journal}, 44(10):2245--2269, 1965.

\bibitem{Held1970}
Michael Held and Richard~M. Karp.
\newblock The traveling-salesman problem and minimum spanning trees.
\newblock {\em Operations Research}, 18(6):1138--1162, 1970.

\bibitem{Helsgaun2009}
Keld Helsgaun.
\newblock General \emph{k}-opt submoves for the lin-kernighan {TSP} heuristic.
\newblock {\em Mathematical Programming Computation}, 1(2-3):119--163, 2009.

\end{thebibliography}

\end{document}